\documentclass[10pt,twocolumn,letterpaper]{article}

\usepackage[pagenumbers]{cvpr} % To force page numbers, e.g. 

\usepackage{graphicx}
\usepackage{amsmath}
\usepackage{amssymb}
\usepackage{booktabs}
\usepackage{float}
\usepackage{booktabs}
\usepackage{multirow}
\usepackage{pifont}
\usepackage[table]{xcolor}
\usepackage{booktabs}
\usepackage{graphicx}
\usepackage{multirow}
\usepackage{float} 
\definecolor{cvprblue}{rgb}{0.21,0.49,0.74}
\usepackage[pagebackref,breaklinks,colorlinks,allcolors=cvprblue]{hyperref}

\title{A Training-Free Guess What Vision Language Model from Snippets to Open-Vocabulary Object Detection}

\author{$\text{Guiying Zhu}^{*1,2}$, $\text{Bowen Yang}^{*2}$, $\text{Yin Zhuang}^{*1,2}$, $\text{Tong Zhang}^{\dag3}$, $\text{Guanqun Wang}^{2}$,\\[2pt] $\text{Zhihao Che}^{1}$, $\text{He Chen}^{1,2}$, $\text{Lianlin Li}^{\dag3}$\\[2pt]
\textsuperscript{1} Aerospace and Informatics Domain, Beijing Institute of Technology, Zhuhai, China\\[2pt]
\textsuperscript{2} National Key Laboratory of Science and Technology on Space-Born Intelligent Information Processing,\\[2pt] School of Information \& Electronics, Beijing Institute of Technology, Beijing, China\\[2pt]
\textsuperscript{3} School of Electronic, Peking University, Beijing, China\\
{\tt\small \{guiying\_z, bowen\_y, yzhuang\}@bit.edu.cn, \{tungz, lianlin.li\}@pku.edu.cn }
}

\begin{document}
\maketitle

\begin{abstract}
Open-Vocabulary Object Detection (OVOD) aims to develop the capability to detect anything. Although myriads of large-scale pre-training efforts have built versatile foundation models that exhibit impressive zero-shot capabilities to facilitate OVOD, the necessity of creating a universal understanding for any object cognition according to already pretrained foundation models is usually overlooked. Therefore, in this paper, a training-free Guess What Vision Language Model, called GW-VLM, is proposed to form a universal understanding paradigm based on our carefully designed Multi-Scale Visual Language Searching (MS-VLS) coupled with Contextual Concept Prompt (CCP) for OVOD. This approach can engage a pre-trained Vision Language Model (VLM) and a Large Language Model (LLM) in the game of \text{“guess what”}. Wherein, MS-VLS leverages multi-scale visual-language soft-alignment for VLM to generate snippets from the results of class-agnostic object detection, while CCP can form the concept of flow referring to MS-VLS and then make LLM understand snippets for OVOD. Finally, the extensive experiments are carried out on natural and remote sensing datasets, including COCO val, Pascal VOC, DIOR, and NWPU-10, and the results indicate that our proposed GW-VLM can achieve superior OVOD performance compared to the-state-of-the-art methods without any training step.
\end{abstract}

\begin{figure*}[!t]
    \centerline{\includegraphics[width=1\textwidth]{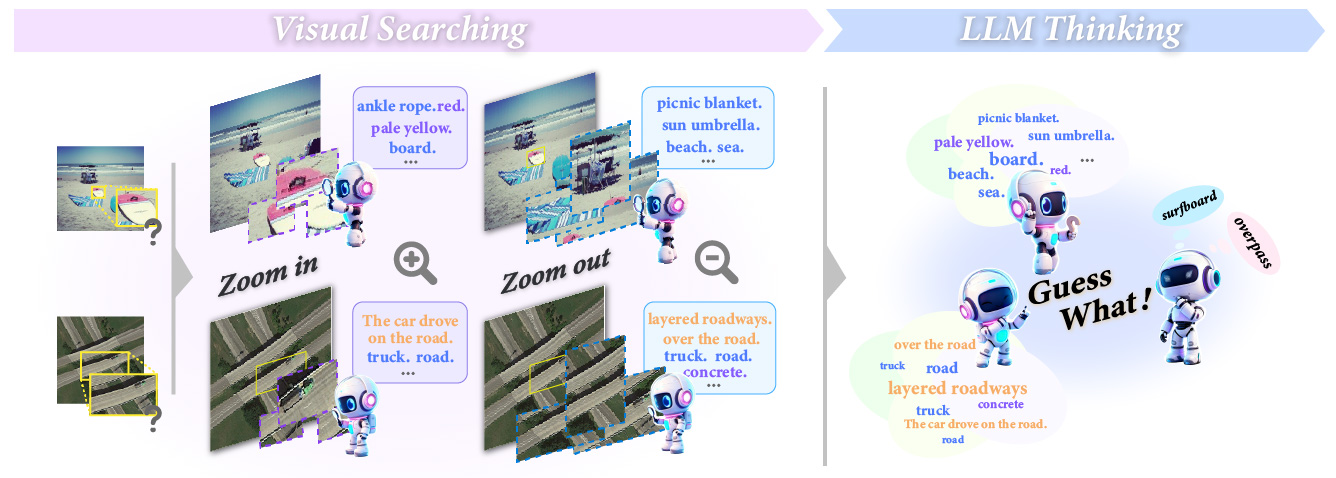}}
    \caption{The pipeline of our \text{‘}Guess What\text{’} game: To understand class-agnostic object based on snippets for OVOD.}
    \label{fig:concept}
\end{figure*}

\section{Introduction}
\label{sec:intro}
\renewcommand{\thefootnote}{} 
\footnote{\textsuperscript{*} Equal Contribution. \textsuperscript{\dag} Corresponding Author.}
Recently, Open-Vocabulary Object Detection (OVOD) has become a new challenge in computer vision field. It is different from traditional object detection methods \cite{ren2015faster,redmon2016you,wang2021fsod,zhang2025unified} that detect objects according to pre-defined closed-set categories, thus aiming to develop the capability to detect anything in various scenarios. Currently, many efforts have been made to improve the generalization ability of object detection, making it more suitable for real-world applications. For example, GLIP \cite{li2022grounded} proves that an object-level, language-aware, and rich visual semantic representation is very important for OVOD. Subsequently, they construct massive image-text data for GLIP pre-training, and then OVOD can be achieved by reformulating object detection as phrase grounding and scaling up visual concepts based on region-text alignment. In view of this, DetCLIPv2 \cite{yao2023detclipv2}, YOLO-World \cite{cheng2024yolo}, and Grounding DINO \cite{liu2024grounding} all focus on introducing diverse datasets for pre-training, including detection, grounding, image-text pairing, and captioning, to establish a better visual semantic representation for OVOD based on their specific region-text alignment. Besides, DetCLIPv3 \cite{yao2024detclipv3}, LLMDet \cite{fu2025llmdet}, and Rex-Omni \cite{jiang2025detect} develop the high information density and large-scale datasets (i.e., 50, 1, and 22 million) for large-scale pre-training or design specific pre-training tasks to further facilitate the improvement of OVOD ability. Although the aforementioned studies can yield improvements in OVOD performance, expensive training resources and labor-intensive efforts for training data preparation and verification are required, which pose burdens for most researchers.

\par Significantly, during the era of large-scale pre-training, it is essential to set up numerous versatile foundation models and effectively utilize their impressive zero-shot capability to facilitate OVOD. Therefore, at present, a large number of versatile foundation models (e.g., SAM \cite{kirillov2023segment}, CLIP \cite{radford2021learning}, DINO \cite{caron2021emerging}, Llama \cite{dubey2024llama}, Qwen \cite{bai2025qwen2}, and so on) have been constructed. However, failing to deliberate on the post-pre-training trajectory and neglecting the underlying capabilities already inherent in foundation models will not only squander substantial computational and human resources but also risk precipitating a protracted stagnation in technological advancement. Therefore, a systematic and in-depth exploration of the capabilities embedded in large-scale pre-trained foundation models is essential for the emergence of advanced intelligence in OVOD. Then, inspired by the studies of InstructSAM \cite{zheng2025instructsam} and the training-free instance segmentation model \cite{espinosa2025no}, this paper proposes a training-free \textbf{G}uess \textbf{W}hat \textbf{V}ision \textbf{L}anguage \textbf{M}odel called GW-VLM to achieve OVOD by establishing a universal understanding paradigm based on already pretrained foundation models. Regarding the formation of a universal understanding paradigm for object cognition, relying solely on isolated region proposals and their individual region–text alignments for OVOD disregards the global spatial dependencies in region proposals. This would severely impair holistic object understanding. Thus, DetCLIPv3 and LLMDet consider hierarchical object labels and global detailed captioning during the pre-training step to introduce comprehensive object-level and image-level information for object cognition in OVOD. Although it can indeed provide human like object understanding information (i.e., object attributions, multi-object relations, and combined meanings of descriptions) to set up semantic concepts assisting in region-text alignment. However, (i) since the \textbf{linguistic descriptive logic exhibits remarkable flexibility} across semantic granularity, relational structure, and frames of reference, it enables dynamic abstraction and uncertainty-aware reasoning while concurrently introducing non-negligible ambiguities that \textbf{necessitate explicit uncertainty modeling or visual feedback for reliable spatial interpretation}. Moreover, (ii) the \textbf{massive detailed description data} involved in captioning, grounding, or multi-label prediction for pre-training \textbf{poses a significant risk of label information leakage}. Inevitably, these two aspects will seriously constrain the ability to understand spatial objects that can truly be oriented towards open vocabulary scenarios.
\par Accordingly, we consider designing Multi-Scale Visual Language Searching (MS-VLS) for GW-VLM to project visual searching information into entangled snippets based on a pre-trained Vision Language Model (VLM) using a soft-alignment technique. Notably, visual searching information is anchored in the results of class-agnostic object detection, which can be seen as visual feedback that assists in understanding class-agnostic objects during the inference phase. Next, a pre-trained Large Language Model (LLM) is considered as a natural open vocabulary library; thus, a Contextual Concept Prompt (CCP) is designed in the proposed GW-VLM to establish the concept of flow for enabling LLM to understand the projected snippets in various open vocabulary scenarios. This can be seen as an interface protocol between the designed MS-VLS and LLM to make the already pre-trained VLM and LLM play the game of  \text{“guess what”} as shown in Figure \ref{fig:concept} to achieve OVOD without any training step. Finally, extensive experiments are carried out on two different OVOD scenarios (i.e., natural and remote sensing scenes) involving four benchmarks (e.g., COCO val, Pascal VOC, DIOR, and NWPU-10), and the results indicate that our proposed GW-VLM can achieve superior OVOD performance compared to the-State-Of-The-Art (SOTA) methods. In general, the contributions of our study can be summarized as follows:
\begin{itemize}
    \item A novel OVOD framework called a training-free GW-VLM is proposed for class-agnostic object understanding, which integrates the already pretrained VLM and LLM at the inference phase to produce visual feedback and form the concept of flow for class-agnostic object cognition in various open vocabulary scenarios.
    
    \item A visual-language searching approach, termed MS-VLS, is proposed to generate visual feedback for class-agnostic objects by leveraging a soft-alignment technique in a pre-trained VLM. This can make the VLM project multi-scale views of class-agnostic objects into entangled snippets, including text, attribution, context, and function phrases, preparing to play a game of \text{“guess what”} with LLM.
    
    \item A concept of flow determined by the CCP is designed to assist LLMs in understanding entangled snippets of class-agnostic objects corresponding to the procedure of MS-VLS, which can enable LLMs to immerse themselves in a fun game of \text{“guess what”} and achieve a more reliable comprehension of OVOD.
\end{itemize}

\section{Related Works}
\textbf{Open Vocabulary Object Detection} aims to establish the ability to detect anything. Consequently, it can be seen as a next generation object detection method that transcends traditional closed-set detectors in the field of computer vision. Hence, many efforts have been made towards OVOD, which can be roughly divided into three types: pre-training, pseudo-labeling, and prompt based methods. For examples, OVR-CNN \cite{zareian2021open}, GLIP \cite{li2022grounded}, OWL-ViT\cite{minderer2022simple}, RO-ViT \cite{kim2023region}, DetCLIPv2 \cite{yao2023detclipv2}, YOLO-World \cite{cheng2024yolo}, Grounding DINO \cite{liu2024grounding}, RTGen \cite{chen2024rtgen}, DetCLIPv3 \cite{yao2024detclipv3}, LLMDet \cite{fu2025llmdet}, Rex-Omni \cite{jiang2025detect}, LLaMA-Unidetector \cite{xie2025llama}, and LAE \cite{pan2025locate} are pre-training-based methods that primarily focus on developing various large-scale multi-modal datasets with specific pre-training strategies to build OVOD ability in downstream tasks. At the same time, pseudo-labeling-based methods, such as OmDet-Turbo \cite{zhao2024real}, OWL-ST \cite{minderer2023scaling}, OW-OVD \cite{xi2025ow}, Object-Centric OVD \cite{bangalath2022bridging}, ViLD \cite{gu2021open}, Dynamic-DINO \cite{lu2025dynamic}, LP-OVOD \cite{pham2024lp}, and CastDet \cite{li2024exploiting}, are employed to leverage the zero-shot ability of pretrained foundation models for OVOD. Finally, the prompt based methods, including DVDet \cite{jin2024llms}, DetPro \cite{du2022learning}, PromptDet \cite{feng2022promptdet}, LBP \cite{li2024learning}, DINO-X \cite{ren2024dino}, OpenRSD \cite{huang2025openrsd}, and LAOD \cite{mumcu2025llm}, provide innovative training or tuning ways by referencing natural language processing (NLP) techniques, which can evidently assist the model in setting up visual concepts for OVOD. Recently, to further reduce computational costs and establish the capabilities of OVOD in the era of large-scale pre-training, InstructSAM \cite{zheng2025instructsam} and the training-free instance segmentation model \cite{espinosa2025no} have begun exploring a training-free OVOD framework based on versatile foundation models.

\par \textbf{The Vision Language Model} is very critical for OVOD to achieve visual region and text phrase alignment. Thereby, the milestone studies, e.g., CLIP \cite{radford2021learning}, ALIGN \cite{jia2021scaling}, BLIP \cite{li2022blip}, and LiT \cite{zhai2022lit}, are widely utilized in the OVOD framework. Nevertheless, the milestone studies can only provide image-text alignment capabilities, which exhibit biases concerning the object-level region-text alignment requirements, thus constraining further performance improvements. Then, LaMI-DERT \cite{du2024lami} designs language model instructions to set up inter-category relations for refining concept representation of isolated category names in the text space of CLIP. OvarNet \cite{chen2023ovarnet} introduces the attribute classification for CLIP, termed CLIP-Attr, to align visual representations with attributes, thus assisting in object-level region-text alignment. CoDet \cite{ma2023codet} groups the object-level images that share a concept in their captions to align co-occurring objects with the shared concept. Next, BARON \cite{wu2023aligning} aligns the embeddings of a bag of regions beyond individual regions to explore the compositional structure of semantic concepts. In summary, these studies of VLM in the OVOD task mainly focus on improving their object-level region-text alignment for understanding class-agnostic objects. 

\par \textbf{Prompt Engineering} is a very important area of study in NLP for probing knowledge from large-scale pre-trained language models. Simultaneously, \cite{ekin2023prompt}, \cite{ye2024prompt}, \cite{schulhoff2024prompt}, and \cite{wang2024prompt} indicate that it is crucial to understand and master the art of prompt engineering to fully harness the potential of LLM in accomplishing complex reasoning tasks. Similarly, several studies \cite{zhou2022learning}, \cite{ren2023prompt}, \cite{chen2025unleashing} focus on designing specific prompting ways to improve the performance of VLM. Recently, with the development of prompt engineering, context engineering \cite{mei2025survey}, \cite{zhang2025agentic}, \cite{hua2025context}, and \cite{li2025c} have been emerged, which is the delicate art and science of filling the context window with just the right information. Consequently, it can enable LLM to understand our situations and purposes in order to elicit expected solutions from LLM without the need for pre-training or fine-tuning additional parameters.

\begin{figure*}[!t]
    \centerline{\includegraphics[width=1.018\textwidth]{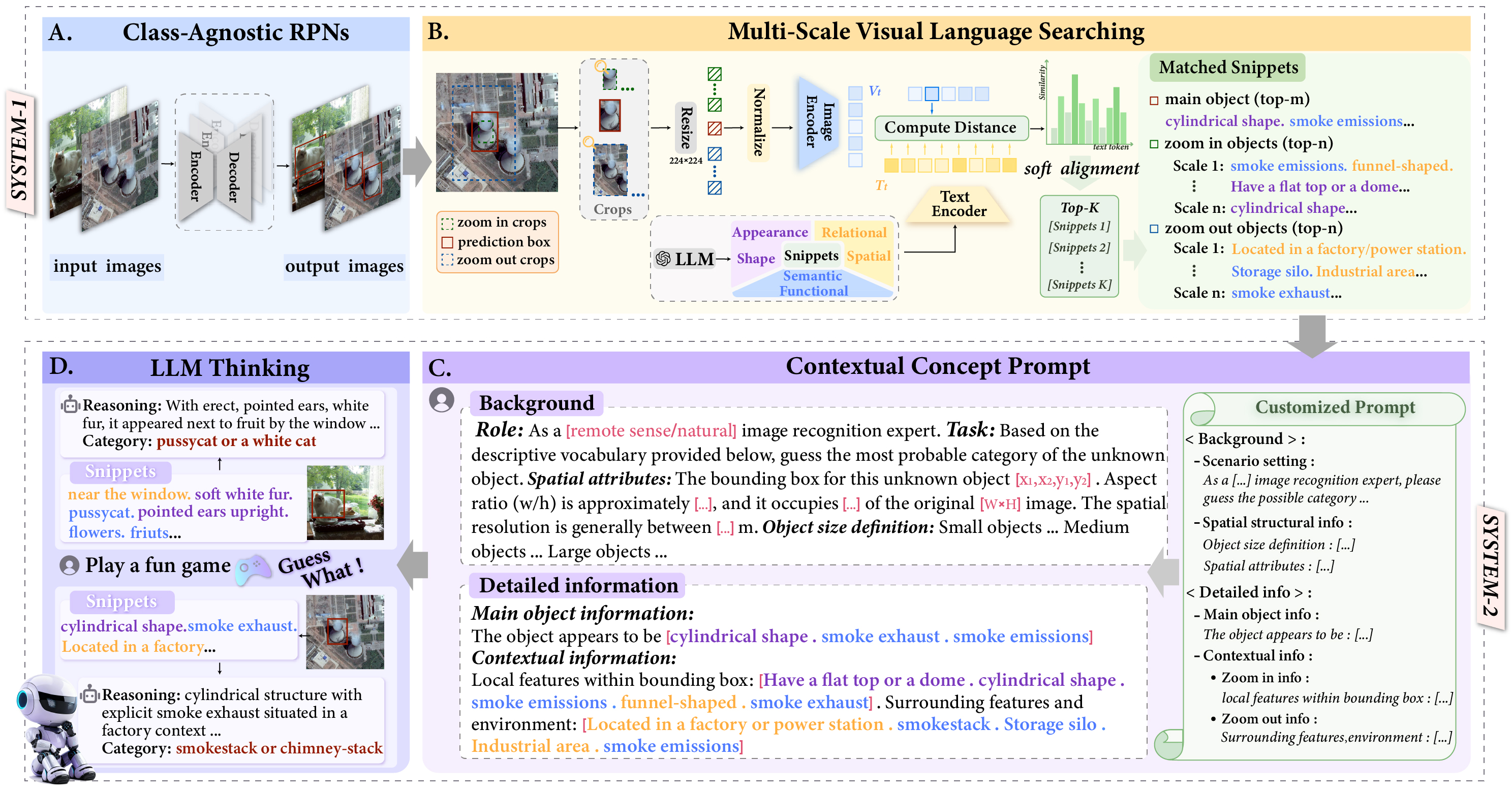}}
    \caption{Overview of GW-VLM. (A) By merging Class-Agnostic RPNs, capturing class-agnostic objects from the input images. (B) Based on the Multi-Scale Visual Language Searching, the extracted crops are soft-aligned with the text tokens in VLM to generate snippets by selecting Top-K semantically matched phrases. (C) The Top-K matched snippets are subsequently embedded into the Contextual Concept Prompt (CCP), which incorporates background information and visual searching information. (D) LLM conducts semantic reasoning and prediction using the CCP.}
    \label{fig:model}
\end{figure*}

\section{Methodology}
\par Referring to the proposed GW-VLM, our main idea is inspired by System 1 and System 2, which have been outlined in \cite{ekin2023prompt}. Especially, System 1 typically requires quick, intuitive, or pattern recognition based answers; thus, in Figure \ref{fig:model} A. and B., there are Class-Agnostic Region Proposal Networks (CA-RPNs) and MS-VLS designed in the proposed GW-VLM to form an innovative pattern recognition method for OVOD, while generating snippets based on a pretrained VLM. System 2 requires the more deliberate, analytical, or complex issue-solving; as a result, in Figure \ref{fig:model} C. and D., a CCP is well designed in the proposed GW-VLM for LLM, which can synthesize the generated snippets from MS-VLS to set up a concept of flow that engages VLM and LLM in the game of \text{“guess what”}, to understand any class-agnostic object and achieve OVOD. Subsequently, the designed CA-RPNs, MS-VLS, and CCP in the proposed GW-VLM are elaborated as follows.

\subsection{Class-Agnostic RPNs} 
\label{subsec:CAR}
As demonstrated by previous studies \cite{zhou2022detecting}, \cite{kirillov2023segment}, and \cite{oquab2023dinov2}, large-scale pre-training can endow an open-vocabulary ability for object-centric interpretation. Therefore, in our study, the large-scale pre-training of RPNs is necessary to achieve OVOD. Fortunately, at present, there are many RPNs that have been pre-trained to detect anything, relying on class-agnostic foreground-background discrimination. As shown in Figure \ref{fig:model} A., a series of class-agnostic RPNs can be configured for the proposed GW-VLM to capture entire class-agnostic objects from images by merging their proposal results, which can be expressed as:
\begin{equation}
   BBoxes=Merge\left\{ RPN_{i}\left( X \right) |i=1,\ldots ,m \right\}.
  \label{eq:1}
\end{equation}
In (\ref{eq:1}), $BBoxes=\left\{ bbox_{CA}^{1},\ldots ,bbox_{CA}^{n} \right\}$ indicates all of detected class-agnostic objects from the input images $X$. Moreover, $Merge\left\{ \cdot \right\}$ indicates the fusion of class-agnostic detection results from $RPN_{i}\left( X \right)$ by using Non-Maximum Suppression (NMS). After that, the configurable RPNs propose sufficient class-agnostic objects preparing for OVOD.

\subsection{Multi-Scale Visual Language Searching}
Referring to these class-agnostic detection results $BBoxes$, establishing a universal understanding paradigm for class-agnostic object cognition becomes a very important issue for OVOD. Now, several studies have investigated object-level region–text alignment through VLMs to enable class-agnostic object understanding. Yet, empirical evidence indicates that the performance of this alignment deteriorates markedly as the open vocabulary codebook expands, even when the model is pre-trained or fine-tuned on extremely large-scale multimodal datasets, due to biases, long-tail distributions, or unpredictable domain shifts. As a result, pursuing an accurate object-level region-text alignment under open-vocabulary scenarios seems impossible. At the same time, SmartCLIP \cite{xie2025smartclip} points out that the pretrained VLMs often struggle with potential information misalignment in many image-text datasets, while suffering from entangled representations and lacking attention to the expected visual main body.
\par However, we consider this misalignment caused by entangled representations to be a very clever phenomenon that perfectly matches the requirements of System 1 (i.e., quick, intuitive, or pattern recognition based answers). It can make already pre-trained VLMs unconcernedly provide entangled snippets (e.g., phrases, texts, descriptions, and so on). Although there are errors or misalignments that occur in these straightforward snippets, In our opinion, once a sufficient number of fragmentary yet correct clues have been accumulated, they can provide reliable support for classification or decision-making. Accordingly, as shown in Figure \ref{fig:model} B., a novel MS-VLS is designed in the proposed GW-VLM to project visual searching information into entangled snippets as visual feedback, preparing for the decision-making of OVOD. First, according to $BBoxes$, multi-scale views are cropped based on each detected class-agnostic anchor, which can be expressed as:
\begin{equation}
 Crops=\begin{cases}BBoxes&\\ Scale_{zoom\ in}^{i}\left( BBoxes \right)&i=1\ldots n\\ Scale_{zoom\ out}^{j}\left( BBoxes \right)&j=1\ldots n\end{cases}
 \label{eq:2}
\end{equation}
In (\ref{eq:2}), $BBoxes$ contain each main body of the prediction bbox, as shown in Figure \ref{fig:model} B., and it serves as a red anchor box for the designed MS-VLS. Subsequently, the entangled multi-scale visual crops $Scale_{zoom\  in}^{i}\left( BBoxes \right)$ and $Scale_{zoom\  out}^{j}\left( BBoxes \right)$ can be captured, as shown in the green and blue dashed anchors of Figure \ref{fig:model} B.. Next, all of the crops are resized into $224\times 224$ and projected as visual tokens $V_{t}$ by the visual encoder from the already pre-trained VLM, respectively. Meanwhile, referring to \cite{garosi2025compositional}, we also randomly sample some appearance, shape, relational, spatial, and semantic functional phrases that are generated from the LLMs (i.e., Llama or Qwen) to enrich the snippets of codebook in the text encoder. Subsequently, all of the snippets in the cookbook are also projected as text tokens $T_{t}$ by the text encoder from the already pre-trained VLM. Significantly, the already pre-trained VLM actually has the ability for visual-language alignment, which can be expressed as:
\begin{equation}
 f_{VLM}\left( Crops,Snippets \right) =\frac{f_{v}\left( V_{t} \right)^{T} f_{t}\left( T_{t} \right)}{\| f_{v}\left( V_{t} \right) \| \cdot \| f_{t}\left( T_{t} \right) \|}.
 \label{eq:3}
\end{equation}
In (\ref{eq:3}), $f_{VLM}\left(\cdot \right)$ indicates the pre-trained visual-language alignment models (e.g., CLIP, BLIP, ALIGN, LiT, and so on). Then, inspired by \cite{yang2025prompt} and \cite{mazzucco2025lost}, we configure the alignment ability from one of these pre-training models to set up visual-language soft-alignment and then project each scale view of visual tokens $V_{t}^{i,j}$ into several entangled snippets of text tokens $T_{t}^{i,j}$. This approach can be formulated as:
\begin{equation}
 f_{Sim}\left( V_{t}^{i,j},T_{t} \right) =\frac{f_{v}\left( V_{t}^{i,j} \right)^{T} f_{t}\left( T_{t} \right)}{\| f_{v}\left( V_{t}^{i,j} \right) \| \cdot \| f_{t}\left( T_{t} \right) \|}.
 \label{eq:4}
\end{equation}
\begin{equation}
 Top-K^{max}\left( f_{Sim}\left( V_{t}^{i,j},T_{t} \right) |f_{Sim}\left( BBoxes,T_{t} \right) \right).
 \label{eq:5}
\end{equation}
In (\ref{eq:4}), $f_{Sim}\left( V_{t}^{i,j},T_{t} \right)$ is used to compute distances by the cosine similarity between each scale visual token $V_{t}^{i,j}$ and all snippet tokens $T_{t}$. 
Then, following (\ref{eq:5}), $Top-K^{max}\left( \cdot \right)$ is a soft-alignment function that projects each scale of MS-VLS into top $K$ similarity snippets conditioned on the main object projection. As shown in Figure \ref{fig:model} B., when followed by our designed MS-VLS, hierarchical visual searching information can be gracefully projected into several matched snippets, entangling the main object by zooming in and out on object scales, thus generating visual clues as visual feedback.

\subsection{Contextual Concept Prompt}
Although multimodal contrastive learning has demonstrated promising open vocabulary capabilities, its reliance on image–text alignment remains superficial and inefficient for understanding class-agnostic concepts in open vocabulary scenarios. Hence, substantial task-specific data and elaborate fine-tuning strategies are still indispensable for adapting these generic representations to different downstream tasks. This situation, much like the distinction between rote memorization and genuine comprehension in human cognition, resembles the intuitive recollection of a concept rather than true understanding. Nevertheless, regarding the OVOD task, it truly requires a model to possess an understanding capability in dealing with class-agnostic object cognition in various open vocabulary scenarios.
\par Accordingly, in our study, we consider that LLMs can serve as a natural open vocabulary library for OVOD, which can be utilized to understand class-agnostic objects from the visual feedback of snippets. Naturally, the core issue turns to endowing LLMs with class-agnostic comprehension capabilities for open vocabulary scenarios. Subsequently, referring to \cite{li2025implicit}, the understanding process can be seen as two steps, including the generation of an internal reasoning trace $z_{1:M}$ and blurting out the expected answer $ans$. First, regarding each LLM $\pi_{\theta}$ given input $Top-K$ snippets for reasoning, which can be expressed as:
\begin{equation}
 z_{1:M}=\left( z_{1},\ldots ,z_{M} \right).
 \label{eq:6}
\end{equation}
\begin{equation}
 \begin{gathered}z_{1:M}\sim \pi_{\theta} \left( \cdot |Top-K \right),\\ ans\sim \pi_{\theta} \left( \cdot |Top-K,z_{1:M} \right).\end{gathered}
 \label{eq:7}
\end{equation}
In (\ref{eq:6}), the internal reasoning trace $z_{1:M}$ can be decomposed into $M$ steps or components that follow a specific logic for comprehension. Then, in (\ref{eq:7}), the expected open vocabulary answer can be acquired, conditioned on $Top-K$ snippets and a specific internal reasoning trace $z_{1:M}$. Thus, as shown in Figure \ref{fig:model} C., a CCP is designed in the proposed GW-VLM to be context engineering, which is developed from prompt engineering to formulate the internal reasoning trace $z_{1:M}$ then set up the communication between System 1 (i.e., CA-RPNs and MS-VLS) and System 2 (CCP and LLM). At the same time, the customized prompt is structured as scenario setting, spatial structural information, main object information, and contextual information for zooming in and out, because \cite{schulhoff2024prompt} indicates that the structure of a prompt, which incorporates conditional or branching logic, can facilitate a more reliable output from LLMs, even comparable with fine-tuning based methods \cite{zhang2025geo} and \cite{fiaz2025geovlm}. Besides, \cite{schulhoff2024prompt} also points out that zero-shot prompting techniques, such as role, style, and emotion prompts, are critical elements that can further improve the performance of LLMs. Thus, as shown in Figure \ref{fig:model} D., a fun game of \text{“guess what”} is introduced to form a concept of flow based on the designed MS-VLS coupled with the designed CCP, which can enable the VLM and LLM in the proposed GW-VLM to play a game and provide superior OVOD performance.

\section{Experiments}
\label{sec:formatting}

\begin{table*}
  \centering
  \caption{Performance comparison of various models on DIOR, using Precision, Recall and F1 score for accuracy(\%).}
  \scalebox{0.695}{
  \begin{tabular}{l|c|ccc|ccc|ccc}
    \toprule
    \multirow{2}*{Method} & \multirow{2}*{Publication} & \multicolumn{9}{c}{DIOR}\\
    \cline{3-11}

     &  & R@IoU0.5 & P@IoU0.5 & F1@IoU0.5 & R@IoU0.95 & P@IoU0.95 & F1@IoU0.95 & R@mIoU & P@mIoU & F1@mIoU \\

    \midrule
    Training-free\cite{espinosa2025no} & arxiv 2025 & 39.01 & 28.89 & 33.19 & 9.04 & 5.42 & 6.78 & 30.02 & 20.55 & 24.40 \\
    Rex-omni\cite{jiang2025detect} & arxiv 2025 & 46.47 & 64.24 & 53.93 & 2.00 & 2.77 & 2.33 & 27.00 & 37.33 & 31.34 \\
    LAE\cite{pan2025locate} & AAAI 2025 & \underline{68.45} & 87.38 & \underline{76.76} & 16.80 & 24.82 & 20.04 & \underline{53.78} & 70.16 & \underline{60.89}\\
    GLIP\cite{li2022grounded} & CVPR 2022 & 16.17 & 5.65 & 8.37 & 0.36 & 0.12 & 0.18 & 9.31 & 3.25 & 4.82 \\
    YOLO-World\cite{cheng2024yolo} & CVPR 2024 & 11.55 & 6.57 & 8.38 & 0.70 & 1.85 & 1.02 & 7.29 & 4.62 & 5.65 \\
    Grounding DINO\cite{liu2024grounding} & ECCV 2024 & 18.12 & 0.64 & 1.23 & 0.83 & 0.02 & 0.04 & 10.69 & 0.45 & 0.86 \\
    % GLIP\cite{li2022grounded} & Y & 16.17 & 5.65 & 8.37 & 0.36 & 0.12 & 0.18 & 9.31 & 3.25 & 4.82 \\
    LLaMA-Unidetector\cite{xie2025llama} & TGRS 2025 & 56.80 & 75.31 & 64.76 & 13.35 & 24.27 & 17.28 & 44.00 & 60.32 & 50.88 \\
    InstructSAM\cite{zheng2025instructsam} & NeurIPS 2025 & 17.00 & 24.84 & 20.18 & 1.85 & 4.79 & 2.67 & 11.62 & 17.74 & 14.04 \\
    \rowcolor{gray!20}
    GW-VLM-Llama & - & \textbf{70.85} & \underline{88.00} & \textbf{78.50} & \textbf{19.30} & \underline{32.64} & \textbf{24.26} & \textbf{56.24} & \underline{71.77} & \textbf{63.06} \\
    \rowcolor{gray!20}
    GW-VLM-Qwen & - & 62.95 & \textbf{93.26} & 75.17 & \underline{17.55} & \textbf{37.34} & \underline{23.88} & 50.10 & \textbf{77.29} & 60.80 \\
    \bottomrule
  \end{tabular}
  }
  % \caption{Performance comparison of various models on DIOR, using Precision, Recall and F1 score for accuracy(\%).}
  \label{tab:contrast-dior} 
\end{table*}

\begin{table*}
  \centering
  \caption{Performance comparison of various models on NWPU-10, using Precision, Recall and F1 score for accuracy(\%).}
  \scalebox{0.695}{
  \begin{tabular}{l|c|ccc|ccc|ccc}
    \toprule
    \multirow{2}*{Method} & \multirow{2}*{Publication} & \multicolumn{9}{c}{NWPU-10}\\
    \cline{3-11}

     &  & R@IoU0.5 & P@IoU0.5 & F1@IoU0.5 & R@IoU0.95 & P@IoU0.95 & F1@IoU0.95 & R@mIoU & P@mIoU & F1@mIoU \\

    \midrule
    Training-free\cite{espinosa2025no} & arxiv 2025 & 44.00 & 42.69 & 43.33 & 1.00 & 1.04 & 1.02 & 32.00 & 25.11 & 28.14\\
    Rex-omni\cite{jiang2025detect} & arxiv 2025 & 58.30 & 29.34 & 39.03 & 2.96 & 1.49 & 1.98 & 39.60 & 19.93 & 26.51 \\ 
    LAE\cite{pan2025locate} & AAAI 2025 & \underline{86.70} & 93.16 & 89.82 & 14.50 & 15.78 & 15.12 & 67.64 & 72.25 & 69.87 \\
    GLIP\cite{li2022grounded} & CVPR 2022 & 38.89 & 9.20 & 14.88 & 3.57 & 0.84 & 1.37 & 27.23 & 6.44 & 10.41 \\
    YOLO-World\cite{cheng2024yolo} & CVPR 2024 & 16.40 & 20.29 & 18.14 & 0.50 & 1.38 & 0.73 & 11.19 & 14.26 & 12.54 \\
    Grounding DINO\cite{liu2024grounding} & ECCV 2024 & 21.98 & 0.45 & 0.88 & 0.48 & 0.01 & 0.02 & 10.81 & 0.22 & 0.43 \\
    % GLIP\cite{li2022grounded} & Y & 38.89 & 9.20 & 14.88 & 3.57 & 0.84 & 1.37 & 27.23 & 6.44 & 10.41 \\
    % InstructSAM\cite{zheng2025instructsam} & NeurIPS 2025 & 36.00 & 47.22 & 40.85 & 1.00 & 2.06 & 1.35 & 24.52 & 33.33 & 28.25 \\
    LLaMA-Unidetector\cite{xie2025llama} & TGRS 2025 & 69.60 & 75.81 & 72.57 & 10.60 & 14.44 & 12.22 & 50.84 & 54.44 & 52.58\\
    InstructSAM\cite{zheng2025instructsam} & NeurIPS 2025 & 36.00 & 47.22 & 40.85 & 1.00 & 2.06 & 1.35 & 24.52 & 33.33 & 28.25 \\
    \rowcolor{gray!20}
    GW-VLM-Llama & - & \textbf{91.70} & \underline{93.23} & \textbf{92.46} & \textbf{20.00} & \underline{23.21} & \textbf{21.74} & \textbf{75.81} & \underline{79.06} & \textbf{77.40} \\
    \rowcolor{gray!20}
    GW-VLM-Qwen & - & 85.80 & \textbf{97.07} & \underline{91.09} & \underline{19.30} & \textbf{24.31} & \underline{21.52} & \underline{71.14} & \textbf{82.08} & \underline{76.22} \\
    \bottomrule

  \end{tabular}
  }
  % \caption{Performance comparison of various models on NWPU-10, using Precision, Recall and F1 score for accuracy(\%).}
  \label{tab:contrast-nwpu}
  
\end{table*}

\begin{table*}
  \centering
  \caption{Performance comparison of various models on COCO val, using Precision, Recall and F1 score for accuracy(\%). }
  \scalebox{0.71}{
  \begin{tabular}{l|c|ccc|ccc|ccc}
    \toprule
    \multirow{2}*{Method} & \multirow{2}*{Publication} & \multicolumn{9}{c}{COCO}\\
    \cline{3-11}

     &  & R@IoU0.5 & P@IoU0.5 & F1@IoU0.5 & R@IoU0.95 & P@IoU0.95 & F1@IoU0.95 & R@mIoU & P@mIoU & F1@mIoU \\

    \midrule
    LLM-Det\cite{fu2025llmdet} & CVPR 2025 & 53.63 & \textbf{83.42} & \underline{65.28} & \textbf{17.47} & \textbf{27.12} & \textbf{21.25} & 44.51 & \textbf{68.96} & \textbf{54.10} \\
    OWL-Vit2\cite{minderer2023scaling} & NeurIPS 2023 & 36.14 & 56.40 & 44.05 & 6.06 & 9.45 & 7.38 & 27.25 & 42.54 & 33.22 \\
    GLIP\cite{li2022grounded} & CVPR 2022 & \textbf{73.88} & 49.69 & 59.42 & 15.28 & 10.28 & 12.29 & \textbf{58.14} & 39.10 & 46.76 \\
    Training-free\cite{espinosa2025no} & arxiv 2025 & 44.55 & 26.93 & 33.57 & 9.60 & 3.04 & 4.62 & 33.55 & 18.29 & 23.67 \\
    Grounding DINO\cite{liu2024grounding} & ECCV 2024 & 62.72 & 60.81 & 61.75 & \underline{17.00} & 16.53 & 16.76 & 49.94 & 48.46 & 49.18 \\
    Yolo-world\cite{cheng2024yolo} & CVPR 2024 & \underline{66.53} & 58.44 & 62.22 & 17.19 & 16.00 & 16.57 & \underline{52.81} & 46.62 & 49.52 \\
    Rex-omni\cite{jiang2025detect} & arxiv 2025 & 60.68 & \underline{71.05} & \textbf{65.46} & 12.91 & 16.77 & 14.59 & 44.84 & \underline{53.12} & 48.63 \\
    OmDet-Turbo\cite{zhao2024real} & arxiv 2024 & 61.02  & 48.93 & 54.31 & 12.30 & 9.86 & 10.95 & 45.61 & 36.57 & 40.60 \\
    \rowcolor{gray!20}
    GW-VLM-Llama & - & 62.20 & 65.50 & 63.81 & 16.50 & \underline{17.52} & \underline{17.01} & 51.42 & 53.01 & \underline{52.20} \\
    \rowcolor{gray!20}
    GW-VLM-Qwen & - & 61.21 & 64.50 & 62.81 & 15.48 & 16.60 & 16.03 & 50.42 & 52.00 & 51.21 \\
    \bottomrule

  \end{tabular}
  }
  % \caption{Performance comparison of various models on COCO val, using Precision, Recall and F1 score for accuracy(\%). }
  \label{tab:contrastcoco}
  
\end{table*}

\begin{table*}
  \centering
  \caption{Performance comparison of various models on Pascal Voc val, using Precision, Recall and F1 score for accuracy(\%). }
  \scalebox{0.71}{
  \begin{tabular}{l|c|ccc|ccc|ccc}
    \toprule
    \multirow{2}*{Method} & \multirow{2}*{Publication} & \multicolumn{9}{c}{Pascal Voc}\\
    \cline{3-11}

     &  & R@IoU0.5 & P@IoU0.5 & F1@IoU0.5 & R@IoU0.95 & P@IoU0.95 & F1@IoU0.95 & R@mIoU & P@mIoU & F1@mIoU \\

    \midrule
    LLM-Det\cite{fu2025llmdet} & CVPR 2025 & 68.57 & \underline{82.45} & 74.87 & 32.33 & \textbf{38.91} & \textbf{35.31} & \underline{72.12} & 59.87 & 65.42 \\
    OWL-Vit2\cite{minderer2023scaling} & NeurIPS 2023 & 53.60 & 74.07 & 62.20 & 20.29 & 28.04 & 23.54 & 45.86 & 63.38 & 53.21 \\
    GLIP\cite{li2022grounded} & CVPR 2022 & 39.20 & 27.99 & 32.66 & 16.17 & 11.55 & 13.47 & 33.42 & 23.87 & 27.85 \\
    Training-free\cite{espinosa2025no} & arxiv 2025 & 65.00 & 40.00 & 49.52 & 14.00 & 4.70 & 7.04 & 47.00 & 28.00 & 35.09 \\
    Grounding DINO\cite{liu2024grounding} & ECCV 2024 & 42.25 & 60.71 & 49.82 & 22.43 & 32.28 & 26.46 & 54.42 & 37.85 & 44.64 \\
    Yolo-world\cite{cheng2024yolo} & CVPR 2024 & 83.21 & 69.80 & 75.91 & \textbf{35.73} & 29.94 & \underline{32.57} & 71.72 & 60.14 & 65.42 \\
    Rex-omni\cite{jiang2025detect} & arxiv 2025 & \underline{84.60} & 80.43 & \textbf{82.46} & \underline{32.60} & 32.40 & 32.50 & 69.73 & 66.76 & \textbf{68.21} \\
    OmDet-Turbo\cite{zhao2024real} & arxiv 2024 & \textbf{87.78} & 44.38 & 58.96 & 32.04 & 16.20 & 21.52 & \textbf{72.46} & 36.64 & 48.67 \\
    \rowcolor{gray!20}
    GW-VLM-Llama & - & 68.69 & 81.32 & 74.47 & 27.76 & 32.87 & 30.10 & 57.26 & \underline{67.79} & 62.08 \\
    \rowcolor{gray!20}
    GW-VLM-Qwen & - & 73.51 & \textbf{83.12} & \underline{78.02} & 29.33 & \underline{35.21} & 31.99 & 60.91 & \textbf{72.34} & \underline{66.13} \\
    \bottomrule

  \end{tabular}
  }
  % \caption{Performance comparison of various models on Pascal VOC val, using Precision, Recall and F1 score for accuracy(\%).}
  \label{tab:contrastvoc}
  
\end{table*}

\begin{figure*}
    \centerline{\includegraphics[width=1.015\textwidth]{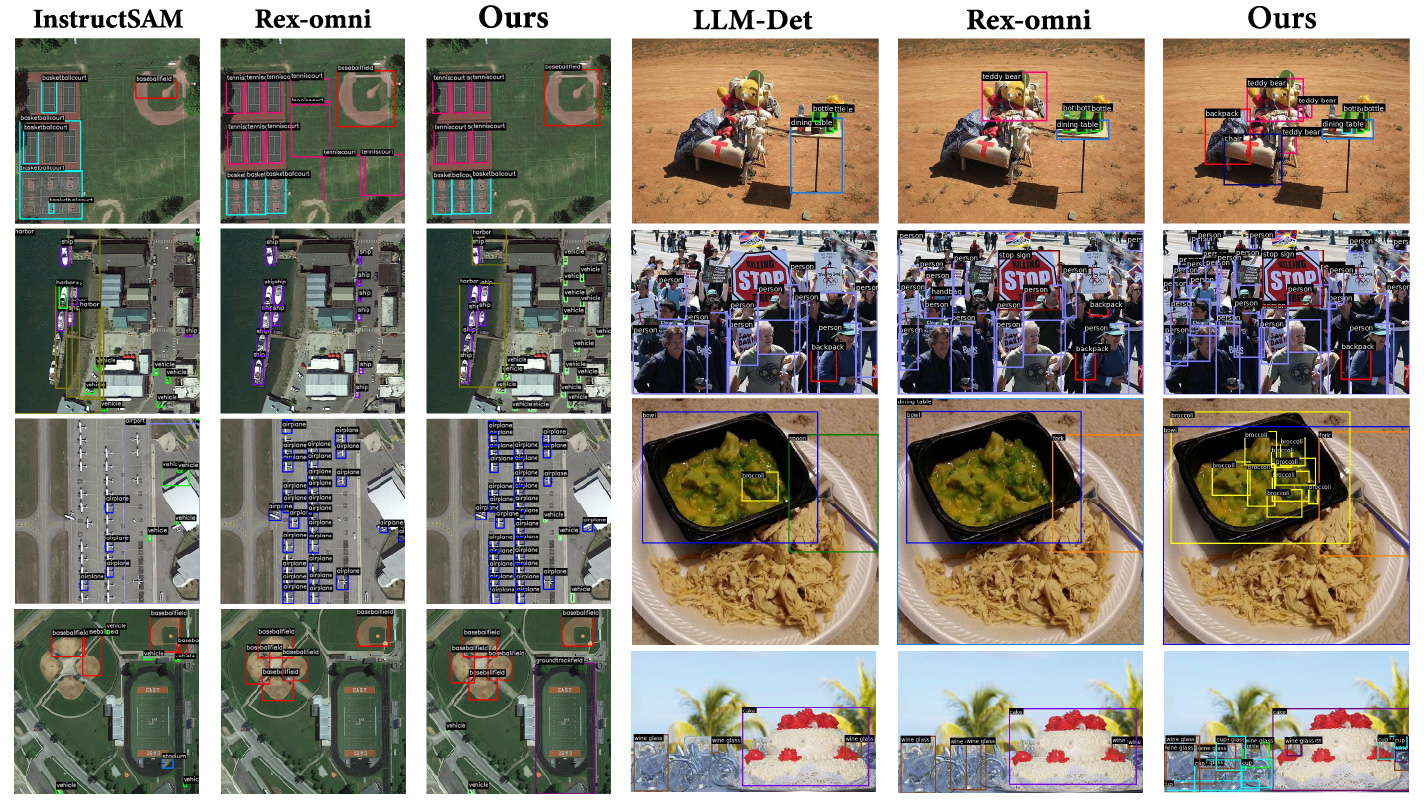}}
    \caption{Visualization of natural and remote sensing objects detection results. Ours refers to the proposed GW-VLM. This figure presents a qualitative comparison of various models on the DIOR and COCO datasets. }
    \label{fig:contrast}
\end{figure*}

\begin{table*}
  \centering
  \caption{Ablation study on NWPU-10 and Pascal VOC using the F1@mIoU score for accuracy(\%). CCP means context concept prompt for LLM. Base-Context means distinguishing only between subject objectives and multi-scale hierarchical contexts. Game-Context refers to detailed task instructions and background information for LLM. Searching-Context incorporates both hierarchical descriptions and higher-order spatial information.}
  \scalebox{0.77}{
  \begin{tabular}{ccc|ccc|c|c}
    \toprule
    \multicolumn{3}{c|}{MS-VLS: RPNs + VLM} & \multicolumn{3}{c|}{CCP: LLM} & \multirow{2}*{NWPU-10} & \multirow{2}*{Pascal VOC} \\
    \cline{1-6}
    Soft-Alignment & Fragmenting-Description & Multi-Scale Searching & Base-Context & Game-Context & Searching-Context & \\

    \midrule
    \ding{51}(Top-1) & & & & & & 35.44 & 47.72 \\
    \ding{51}(Top-K) & \ding{51} & & \ding{51} & & & 69.19 ({\raisebox{0.5pt}{\scriptsize +33.75$\uparrow$}}) & 57.42 ({\raisebox{0.5pt}{\scriptsize +9.70$\uparrow$}})\\
    \ding{51}(Top-K) & \ding{51} & \ding{51} & \ding{51} & & & 74.49 ({\raisebox{0.5pt}{\scriptsize +5.30$\uparrow$}}) & 59.75 ({\raisebox{0.5pt}{\scriptsize +2.33$\uparrow$}})\\
    \ding{51}(Top-K) & \ding{51} & \ding{51} & \ding{51} & \ding{51} & & 76.66 ({\raisebox{0.5pt}{\scriptsize +2.17$\uparrow$}}) & 63.17 ({\raisebox{0.5pt}{\scriptsize +3.42$\uparrow$}})\\
    \ding{51}(Top-K) & \ding{51} & \ding{51} & \ding{51} & \ding{51} & \ding{51} & \textbf{77.40} ({\raisebox{0.5pt}{\scriptsize +0.74$\uparrow$}}) & \textbf{66.13} ({\raisebox{0.5pt}{\scriptsize +2.96$\uparrow$}})\\
    \bottomrule
  \end{tabular}
  }
  % \caption{Ablation study on NWPU-10 and Pascal VOC using the F1@mIoU score for accuracy(\%). CCP means context concept prompt for LLM. Base-Context means distinguishing only between subject objectives and multi-scale hierarchical contexts. Game-Context refers to detailed task instructions and background information for LLM. Searching-Context incorporates both hierarchical descriptions and higher-order spatial information.}
  \label{tab:ablation}
\end{table*}

\begin{table}[htbp] 
  \centering
  \caption{Open vocabulary analysis on DIOR dataset, using F1@IoU0.5 for accuracy(\%).}
  \scalebox{0.80}{
  \begin{tabular}{l|ccc|c}
    \toprule
    \multirow{2}*{Method} & \multicolumn{4}{c}{DIOR}\\
    \cline{2-5}
     & texts-1 & texts-2 & texts-3 & Avg \\
    \midrule
    LAE\cite{pan2025locate} & \underline{62.10} & \underline{56.17} & \underline{54.39} & \underline{57.55}\\
    YoloWorld\cite{cheng2024yolo} & 9.68 & 6.18 & 9.18 & 8.34\\
    Rex-omni\cite{jiang2025detect} & 51.55 & 51.35 & 51.79 & 51.56\\
    \rowcolor{gray!20}
    GW-VLM & \textbf{76.83} & \textbf{76.11} & \textbf{76.22} & \textbf{76.40}\\
    \bottomrule
  \end{tabular}
  }
  % \caption{Open vocabulary analysis on DIOR dataset, using F1@IoU0.5 for accuracy(\%).}
  \label{tab:change_dior} 
\end{table} 

  \bigskip  
  
\begin{table}[htbp]
  \centering
  \caption{Open vocabulary analysis on COCO dataset, using F1@IoU0.5 for accuracy(\%).}
  \scalebox{0.80}{
  \begin{tabular}{l|ccc|c}
    \toprule
    \multirow{2}*{Method} & \multicolumn{4}{c}{COCO}\\
    \cline{2-5}
     & texts-1 & texts-2 & texts-3 & Avg \\
    \midrule
    Grounding-Dino\cite{liu2024grounding}  & 16.63 & 31.14 & 29.62 & 25.79\\
    YoloWorld\cite{cheng2024yolo} & 21.21 & 36.12 & 37.84 & 34.26\\
    LLM-Det\cite{fu2025llmdet} & \underline{27.21} & \underline{45.57} & \underline{45.52} & \underline{39.44}\\
    \rowcolor{gray!20}
    GW-VLM & \textbf{52.17} & \textbf{58.02} & \textbf{57.78} & \textbf{55.99}\\
    \bottomrule
  \end{tabular}
  }
  % \caption{Open vocabulary analysis on COCO dataset, using F1@IoU0.5 for accuracy(\%).}
  \label{tab:change_coco} 
  
\end{table}

\subsection{Implementation Details}
\label{sec:Implementation Details}
To comprehensively evaluate the generalization capability of our proposed GW-VLM, we conduct OVOD testing on four challenging benchmarks spanning both natural scenes and the remote sensing domain, including COCO 2017 val, Pascal VOC 2012, DIOR, and NWPU-10. Meanwhile, the indicators of Precision (P), Recall (R) and F1 are employed. Besides, in order to better compare our proposed OVOD framework of GW-VLM within a closed set, we re-project the open vocabulary results from LLM into the stationary categories involved in the datasets of COCO 2017 val, Pascal-VOC 2012, DIOR, and NWPU-10 for evaluation. 
\par Next, the designed CA-RPNs merge the results of several famous pre-trained RPNs in natural and remote sensing scenes, respectively. Then, the pre-trained VLMs of SigLIP \cite{zhai2023sigmoid} and RemoteCLIP \cite{liu2024remoteclip} are individually configured for the designed MS-VLS to deal with natural and remote sensing scenes. Subsequently, the searching strategy in natural scenes includes the primary view (scale 1.0x), one zoom-out scale, and two zoom-in scales that are adaptively determined based on the sizes of objects. Besides, the soft-alignment at each scale selects the Top-3 snippets. For remote sensing scenes, the targets are categorized into large, medium, and small sizes, and the scaling factors are adaptively chosen based on the target size. At the same time, two to three zoom-out scales and two to three zoom-in scales are selected for visual-language searching. Then, Top-3 snippets are selected for primary view, while Top-5 snippets are selected for the other zoom scales. Here, whether for natural or remote sensing scenes, Qwen-Plus and Llama3.3-70B are equipped for the proposed GW-VLM for a fun game of \text{“guess what”}. Finally, to ensure the deterministic output of LLMs, we set the temperature parameter $\mathbf{T \in [0.0, 0.1]}$.

% \begin{figure*}[!t]
%     \centerline{\includegraphics[width=1.015\textwidth]{Fig/fig-contrast.jpg}}
%     \caption{Visualization of natural and remote sensing objects detection results. Ours refers to the proposed GW-VLM. This figure presents a qualitative comparison of various models on the DIOR and COCO datasets. }
%     \label{fig:contrast}
% \end{figure*}

\subsection{Comparison Results}
\label{sec:Comparison analysis}
We evaluate our GW-VLM against several models, including open-vocabulary detectors and visual-language models, in both natural scenes and 
remote sensing scenes to verify the generalization of our model. 
Our results on the remote sensing benchmarks are reported in Tab.~\ref{tab:contrast-dior} and Tab.~\ref{tab:contrast-nwpu}.
Our model establishes a new SOTA, achieving top F1@mIoU scores of 63.06\% on DIOR and 77.40\% on NWPU-10. This performance significantly exceeds the training-free baseline InstructSAM \cite{zheng2025instructsam}, obtaining a massive gain of 49.02\% F1@mIoU on DIOR and 49.15\% F1@mIoU on NWPU-10. Moreover, our method also outperforms the pretraining baseline LAE \cite{pan2025locate}, exceeding it by 2.17\% F1@mIoU on DIOR and 7.53\% F1@mIoU on NWPU-10. For natural scene benchmarks, our results are reported in Tab.~\ref{tab:contrastcoco} and Tab.~\ref{tab:contrastvoc}.
On these considerable complex benchmarks, our GW-VLM achieves highly competitive results, with 50.20\% F1@mIoU on COCO val and 66.13\% F1@mIoU on PASCAL VOC, and surpasses several models pretrained on large-scale datasets. From the tables regarding 0.95 IoU, merging the results of RPNs that distinguish only between foreground and background exhibits powerful localization ability. For a more intuitive display of the OVOD ability of our proposed GW-VLM, several visualized comparison results on both natural and remote sensing scenes are shown in Figure \ref{fig:contrast}.

% Table\ref{tab:contrast-dior} and table\ref{tab:contrast-nwpu} compare  the results achieved by various OVOD models in the fields of remote sensing and natural domains on DIOR and NWPU-10 datasets. GW-VLM surpasses exiting approaches, 

\subsection{Ablation Study}
\label{sec:Ablation Study}
We perform ablation experiments on NWPU-10 and Pascal VOC datasets to explore the impact of MS-VLS and CCP in the proposed GW-VLM, as reported in Tab.~\ref{tab:ablation}. By incorporating soft-aligned Top-K fragmenting descriptions and base context for LLM, notable improvements can be observed in the performance metrics. Specifically, it achieves a 33.75\% F1@mIoU increase on NWPU, and increases by 9.70\% F1@mIoU on Pascal VOC. This result indicates that LLM hold a very crucial role in open-vocabulary detection tasks, with their significance on par with that of classifiers in closed-set classification or decision-making tasks. Then, following the incorporation of MS-VLS and the continuous refinement of the context prompt for LLM, we observed further improvements, with a significant increase of 8.21\% on NWPU and a corresponding improvement of 8.78\% on Pascal VOC.

\subsection{Open Vocabulary Analysis}
\label{sec:Open Vocabulary Analysis}
Previous OVOD methods are highly sensitive to specific phrasing of the input prompts during training and inference phases. This dependency leads to poor robustness for real OVOD when the pre-defined prompts are changed. For example, if the label is replaced with a semantically equivalent noun that refers to the same visual concept in the real world, the final performance of the models will degrade significantly. This phenomenon indicates a lack of truly open-vocabulary ability, as the models are sensitive to lexical choice rather than the underlying semantic meaning. However, our GW-VLM can overcome this dependency due to the soft alignment mechanism and achieve a truer form of open-vocabulary ability.

\par To validate this, the prompt swapping experiments are conducted on the datasets of COCO val and DIOR. First, we generate a new set of prompts based on the original categories, where each class label is replaced with a semantically equivalent but textually different noun. Subsequently, the new sets of prompts are used for inference. Here, in order to facilitate evaluation, we also re-mapped the new sets of prompts to annotations for the correct metric. For each dataset, three new sets of prompts are generated for evaluation, and all comparison models are subject to the same evaluation protocol, being tested on both these newly generated prompt sets, i.e., text-1, text-2, and text-3. The comparison results are reported in Tab.~\ref{tab:change_dior} and Tab.~\ref{tab:change_coco}, and we can see that our proposed GW-VLM can achieve more stable performance under different prompt swapping experiments, which demonstrates our proposed GW-VLM has a powerful OVOD ability. 

\section{Conclusion}
In this study, a novel training-free framework called GW-VLMs is proposed for OVOD based on integrating the impressive zero-shot abilities of pre-trained RPNs, VLMs, and LLMs. A new MS-VLS and CCP are designed to establish a concept of flow that enables GW-VLM to achieve OVOD by engaging in a fun game of \text{“guess what”} between the pre-trained VLM and LLM, relying on class-agnostic object detection. Finally, extensive testing experiments are carried out, and the results prove that GW-VLM can be a feasible way.

{
    \small
    \bibliographystyle{ieeenat_fullname}
    \bibliography{main}
}

\clearpage
\setcounter{page}{1}
\maketitlesupplementary

% \section{Rationale}
% \label{sec:rationale}
% % 
% Having the supplementary compiled together with the main paper means that:
% % 
% \begin{itemize}
% \item The supplementary can back-reference sections of the main paper, for example, we can refer to \cref{sec:intro};
% \item The main paper can forward reference sub-sections within the supplementary explicitly (e.g. referring to a particular experiment); 
% \item When submitted to arXiv, the supplementary will already included at the end of the paper.
% \end{itemize}
% % 
% To split the supplementary pages from the main paper, you can use \href{https://support.apple.com/en-ca/guide/preview/prvw11793/mac#:~:text=Delete%20a%20page%20from%20a,or%20choose%20Edit%20%3E%20Delete).}{Preview (on macOS)}, \href{https://www.adobe.com/acrobat/how-to/delete-pages-from-pdf.html#:~:text=Choose%20%E2%80%9CTools%E2%80%9D%20%3E%20%E2%80%9COrganize,or%20pages%20from%20the%20file.}{Adobe Acrobat} (on all OSs), as well as \href{https://superuser.com/questions/517986/is-it-possible-to-delete-some-pages-of-a-pdf-document}{command line tools}.

\section*{Acknowledgments}
\label{sec:Acknowledgments}
This work was supported by the General Program of National Natural Science Foundation of China under grant 62371048, in part by the Ye Qisun Science Foundation of the National Natural Science Foundation of China under Grant U2341202 and the Postdoctoral Fellowship Program of CPSF under Grant Number GZC20250393.

\begin{figure}[!htbp]
    \centerline{\includegraphics[width=0.50\textwidth]{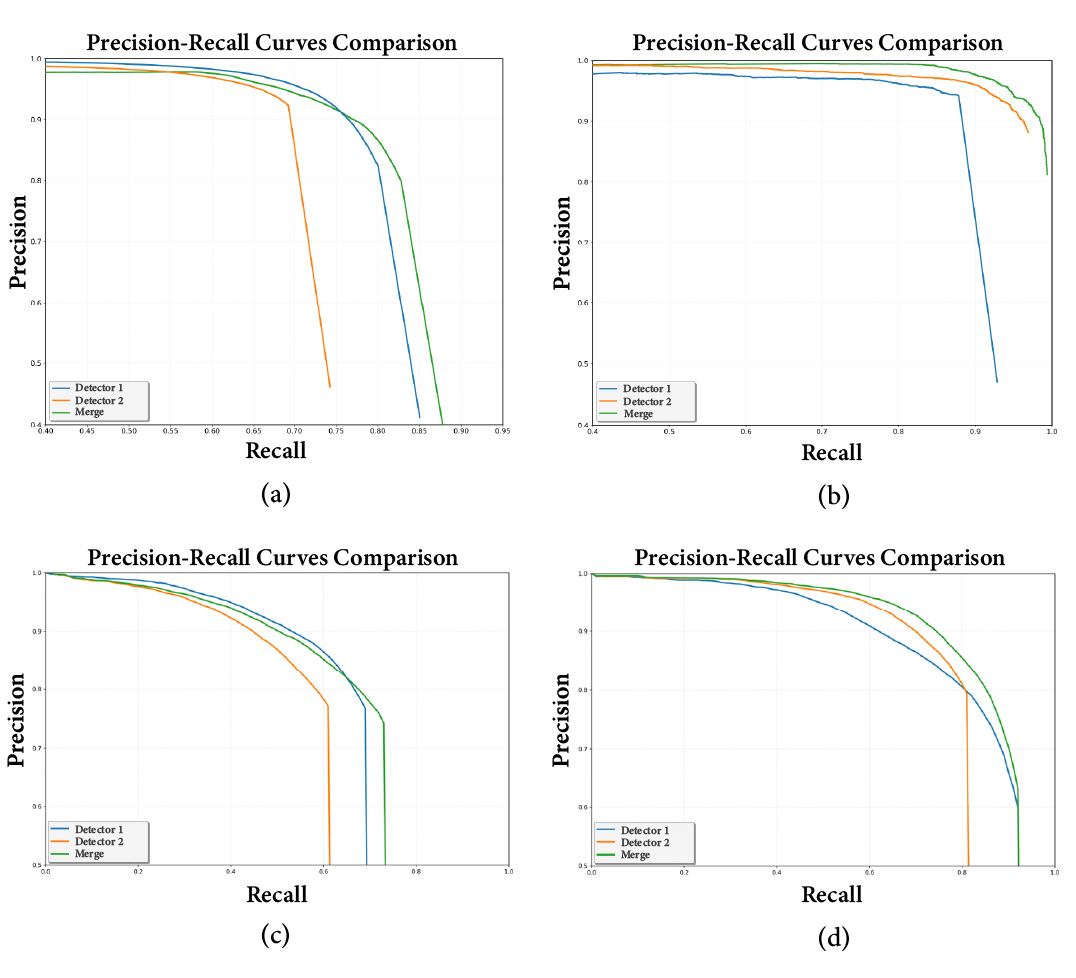}}
    \caption{P-R curves of detector fusion on four datasets: (a) DIOR, (b) NWPU-10, (c) COCO, (d) Pascal VOC.}
    \label{fig:pr}
\end{figure}

\begin{figure*}[t]
    \centerline{\includegraphics[width=1.0\textwidth]{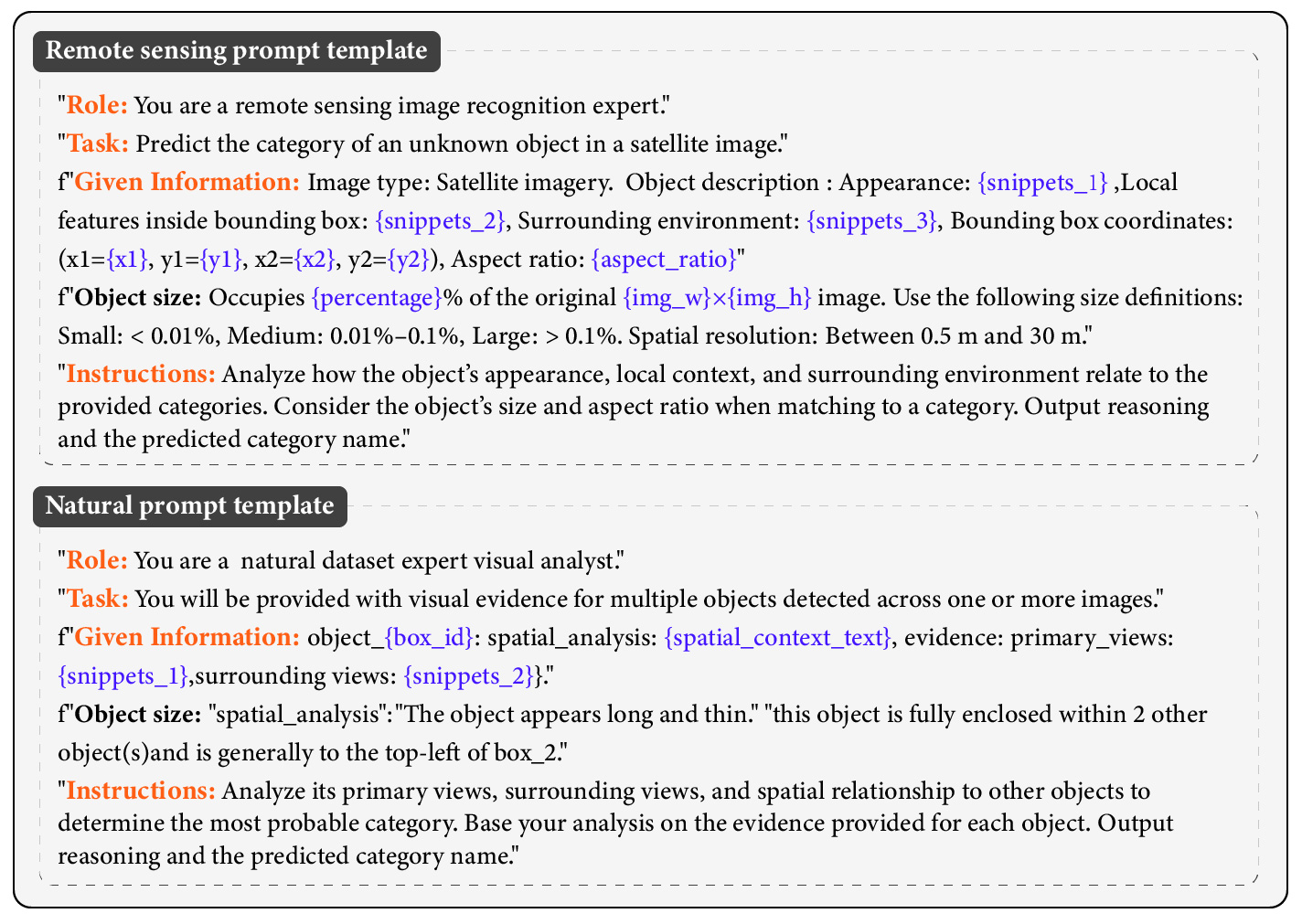}}
    \caption{Prompt templates for both remote sensing and natural scenes}
    \label{fig:prompt}
\end{figure*}

\section*{Effectiveness of Detector Merging Strategy}
\label{sec:merge_analysis}

% As mentioned in \ref{subsec:CAR} of the main paper, we employ a Non-Maximum Suppression (NMS) based fusion strategy to integrate the detection results from different detectors.\ref{fig:pr}
% The visualization results for remote sensing scenes are presented in Figure , while the results for natural scenes are illustrated in Figure .
As introduced in \ref{subsec:CAR} of the main paper, the class-agnostic detection results are integrated using a Non-Maximum Suppression(NMS) based fusion strategy. Then, the P-R curve comparisons on both remote sensing datasets (i.e., DIOR and NWPU-10) and natural scene datasets (i.e., COCO val and Pascal VOC) are illustrated in Figure~\ref{fig:pr}. Specifically, for natural scenes, the fused results of class-agnostic detection from \textbf{DINOv3} and \textbf{LLMDet} indicate a clear gain in True Positives (TP). For example, the merged result achieves \textbf{14,617} TP (14,159 TP for DINOv3 and 12,911 TP for LLMDet, respectively) on Pascal VOC (val), and it also reaches \textbf{27,572} TP (surpassing 25,270 TP for DINOv3 and 23,359 TP for LLMDet) on COCO (val). Similarly, for remote sensing scenes, the fused class-agnostic detection results from \textbf{LAE} and \textbf{Skysense} have \textbf{113,236} TP on DIOR comparing to 102,815 TP for LAE and 69,084 for Skysense, and it also achieves \textbf{3,820} on NWPU-10 comapring to 3,369 TP for LAE and 3,684 TP for Skysense. Furthermore, an extended recall range is observed under the fused setting, reflecting the fusion strategy’s capacity to preserve a substantially larger subset of true objects. Meanwhile, the merged recall ultimately leads to a more exhaustive and complete collection of candidate bounding boxes, effectively reducing missed detections and capturing a broader spectrum of object instances.

\section*{Snippets for VLMs}
\label{sec:snippets}

% As introduced in the main paper, we design the input text snippets to be hierarchical and endowed with rich semantic information. Unlike simple category names, these snippets provide structured context that guides the Vision-Language Model (VLM) to recognize objects more accurately. 

% Specific examples of these generated snippets are presented in Table~\ref{tab:snippets}. The table showcases the prompts constructed for both natural scenes and remote sensing scenes, demonstrating the adaptability and semantic depth of our approach.

The input texts for producing snippets are constructed in a hierarchical fashion and enriched with fine-grained semantic cues. Therefore, in our study, rather than relying solely on simple category names, the designed snippets capture multi-level contextual information, including appearances, shape attributes, spatial, semantic, and functional cues, and so on. Such structured descriptions are provided by Vision-Language Models (VLMs) with more informative guidance, then facilitating more reliable and discriminative open vocabulary object cognition. Notably,
representative examples of these pre-defined snippets are provided in Table~\ref{tab:snippets}. The table reports the prompts formulated for both natural scenes and remote sensing imagery, illustrating the semantic richness, descriptive granularity, and cross-domain adaptability embedded in the snippet construction process.

\section*{Prompt Templates}
\label{sec:templates}
Specific prompt templates for both remote sensing and natural scenes are provided in Figure~\ref{fig:prompt}. About remote sensing scenes, the fixed overhead viewpoint and the lack of familiar perspective cues make it difficult for the model to infer scale from appearance alone. Thereby, the prompt for each remote sensing object explicitly injects information about spatial resolution and object size (e.g., height and width), together with the bounding box extent, aspect ratio, and the fraction of the image area occupied. The template also maps this fraction to discrete size levels (i.e., small, medium, and large) at different resolutions, summarizes the object's appearance, local features inside the bounding box and surrounding environment using obtained snippets. The LLM is instructed to combine these appearance, context, size cues to reason about the category and output both its reasoning process and the final predicted category name. In contrast, for natural scenes, objects are captured from diverse viewpoints with rich contextual interactions. Hence, the prompt for the natural scene focuses on relational and multi-view evidence instead of an explicit metric scale. Then, the template needs to be filled in with snippets of the object based on primary and surrounding views, along with a structural text analysis that characterizes its properties. Rather than specifying resolution and physical size, the natural prompt emphasizes these relative spatial relationships and scene level context, instructing the LLM to use them to determine the most probable category.
% Given the fixed overhead view in remote sensing, we explicitly incorporate resolution and object dimensions (H/W) to address scale variations. In contrast, for natural scenes, we adopt relative spatial relationships to leverage contextual cues.

\section*{Qualitative Analysis of Inference and Reasoning}
\label{Qualitative s}

\begin{figure*}[!t]
    \centerline{\includegraphics[width=0.995\textwidth]{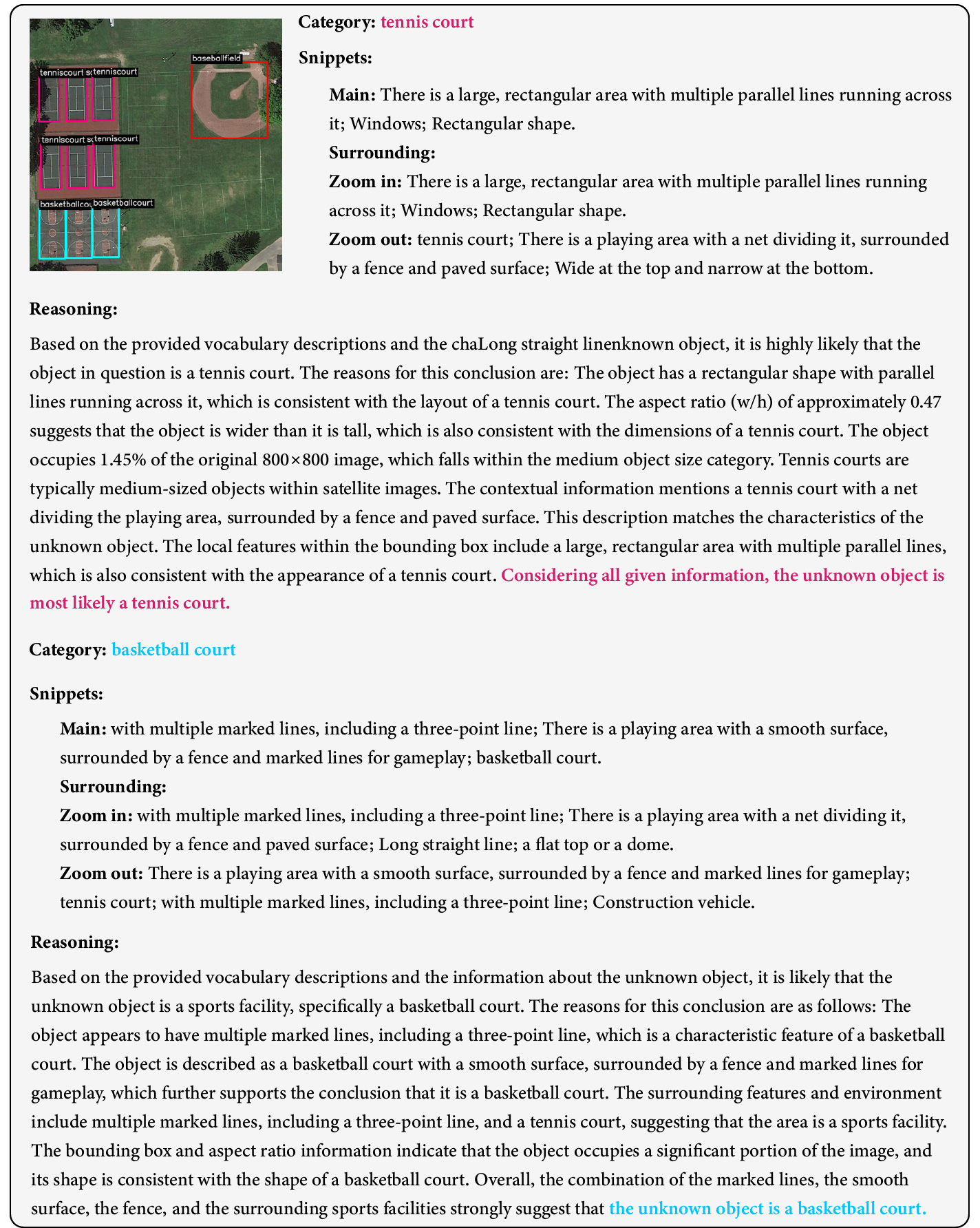}}
    \caption{LLM reasoning with snippet-based prompts to predict the categories ``tennis court" and ``basketball court".}
    \label{fig:rs_1}
\end{figure*}

\begin{figure*}[!t]
    \centerline{\includegraphics[width=0.975\textwidth]{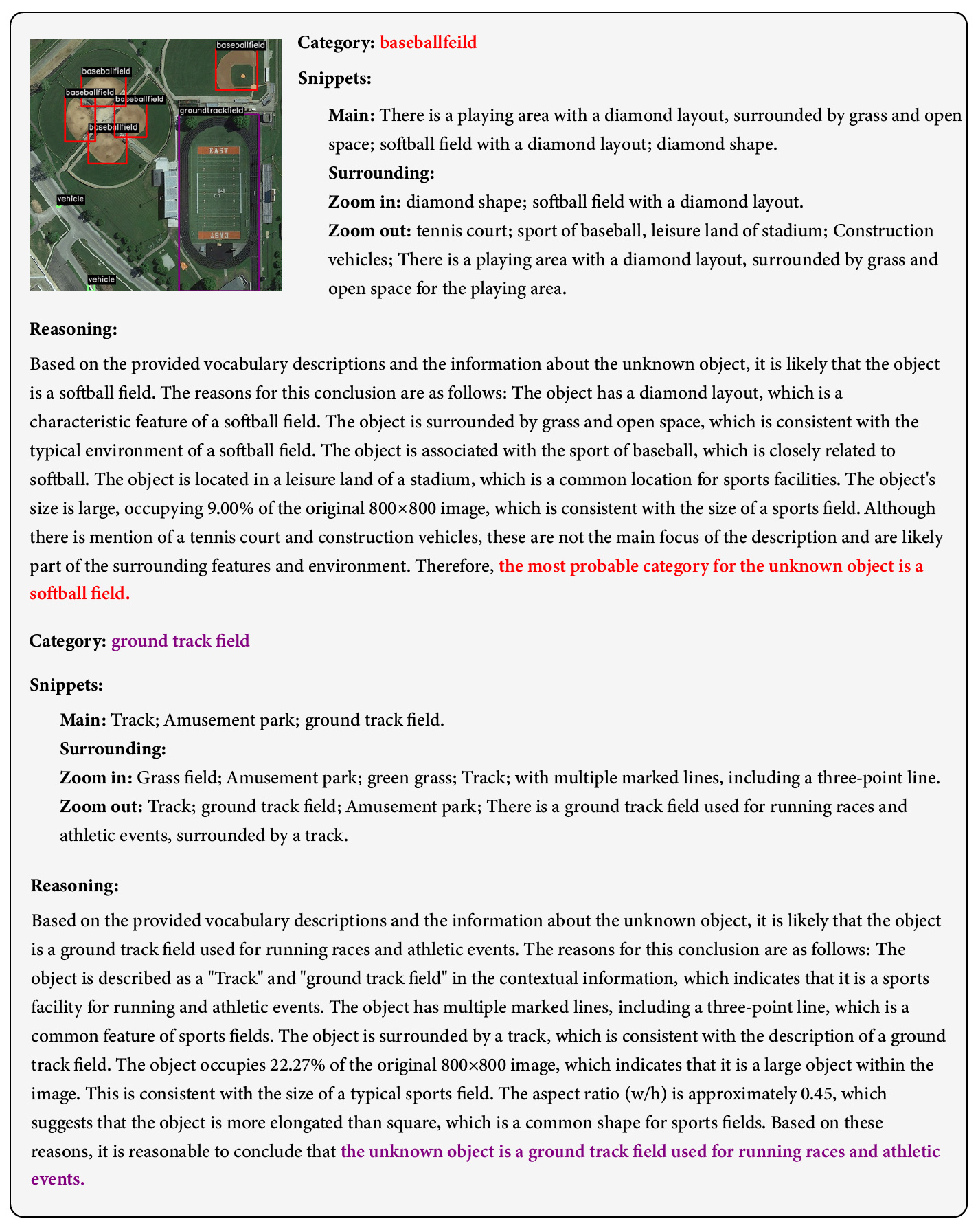}}
    \caption{LLM reasoning with snippet-based prompts to predict the categories ``baseball field" and ``ground track field".}
    \label{fig:rs_2}
\end{figure*}

\begin{figure*}[!t]
    \centerline{\includegraphics[width=0.995\textwidth]{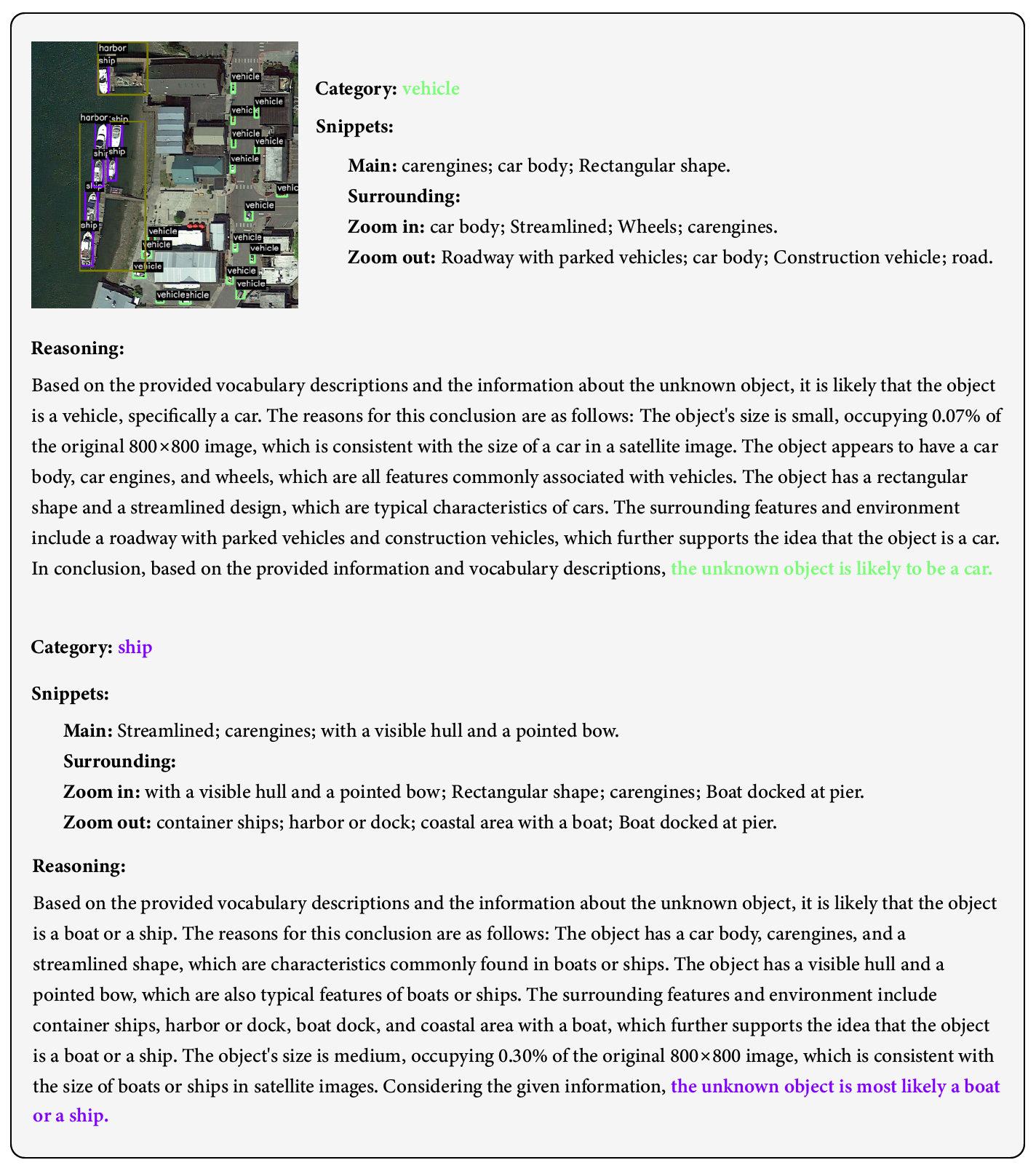}}
    \caption{LLM reasoning with snippet-based prompts to predict the categories ``vehicle" and ``ship".}
    \label{fig:rs_3}
\end{figure*}

\begin{figure*}[!t]
    \centerline{\includegraphics[width=0.995\textwidth]{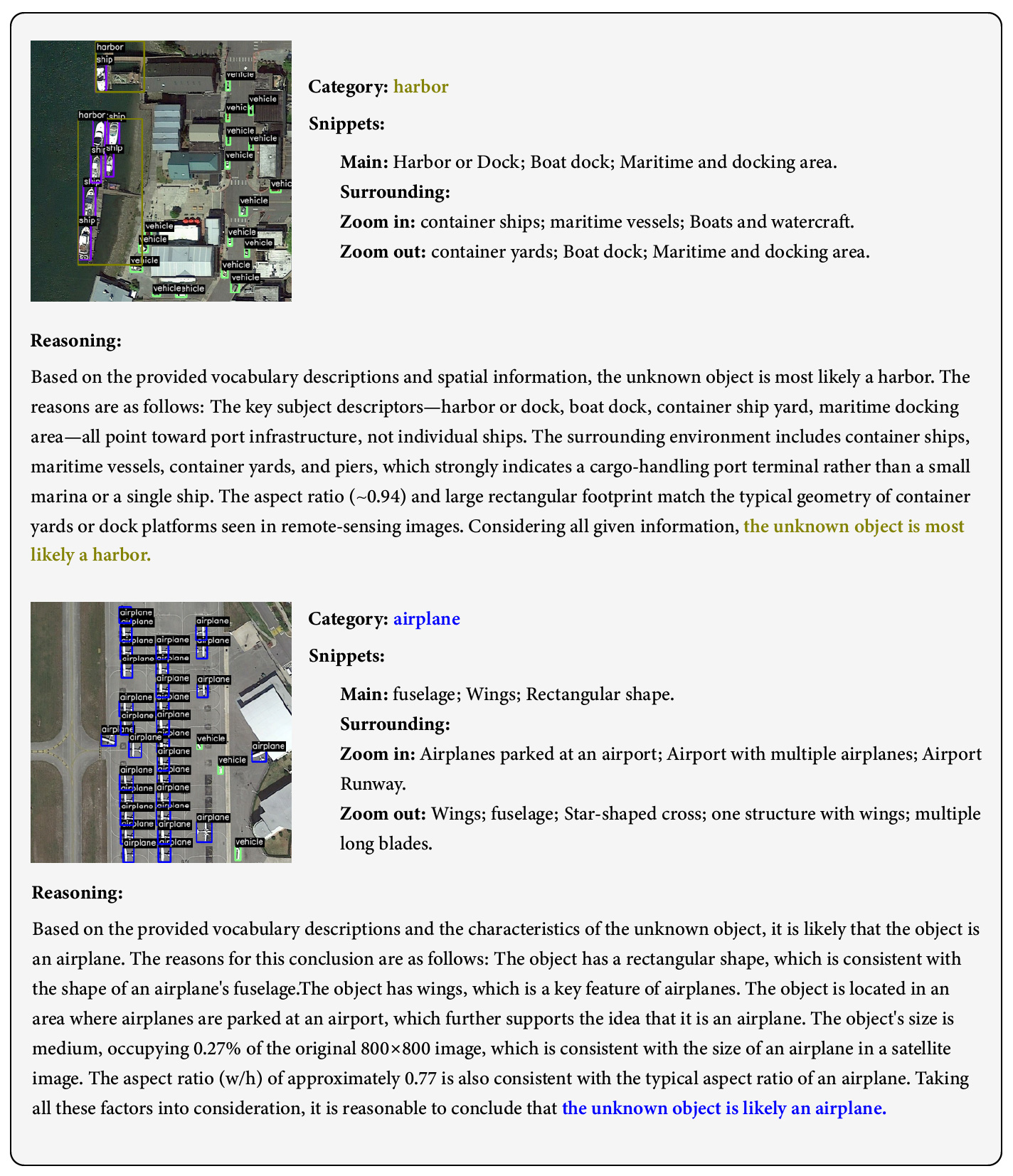}}
    \caption{LLM reasoning with snippet-based prompts to predict the categories ``harbor" and ``airplane".}
    \label{fig:rs_4}
\end{figure*}

\begin{figure*}[!t]
    \centerline{\includegraphics[width=0.975\textwidth]{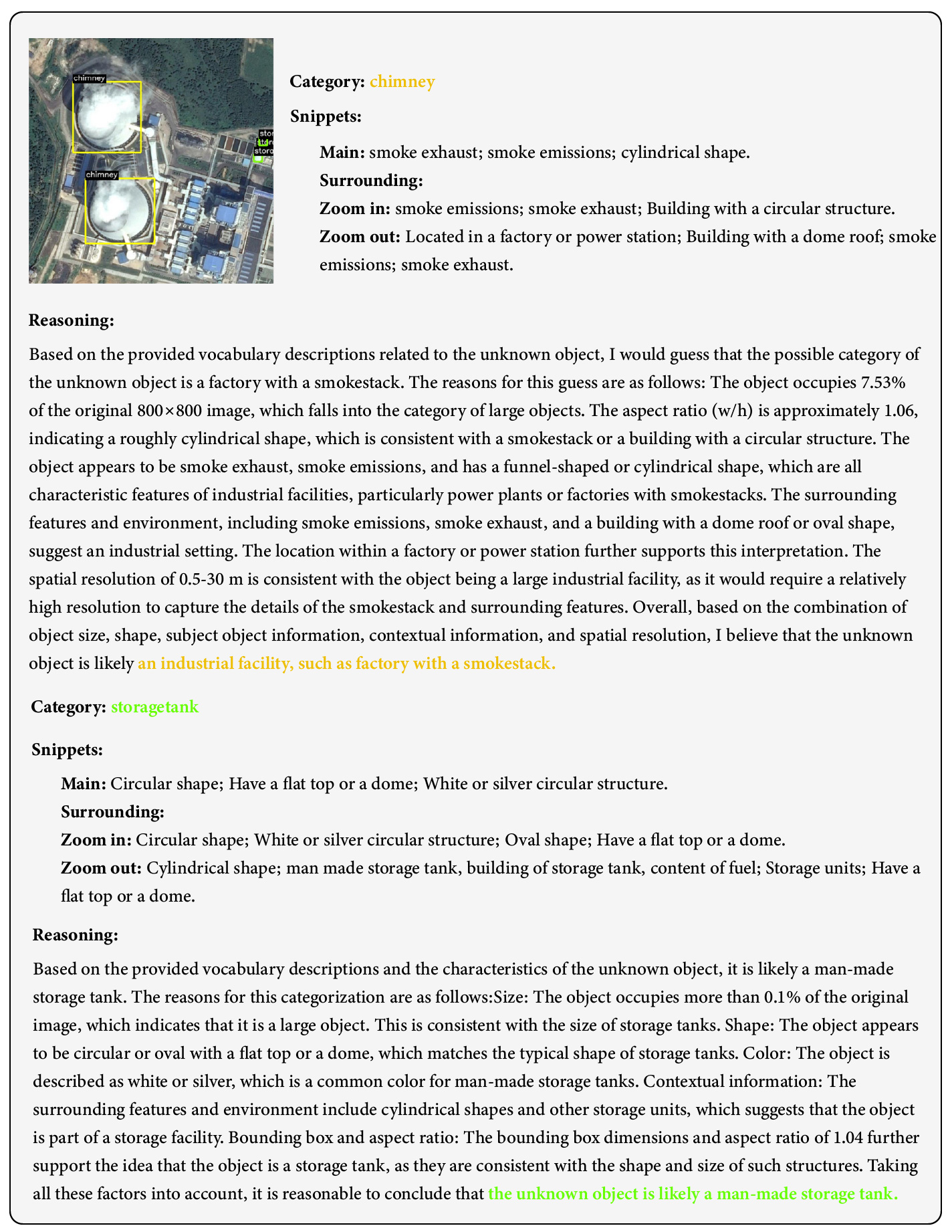}}
    \caption{LLM reasoning with snippet-based prompts to predict the categories ``chimney" and ``storage tank".}
    \label{fig:rs_5}
\end{figure*}

\begin{figure*}[!t]
    \centerline{\includegraphics[width=0.995\textwidth]{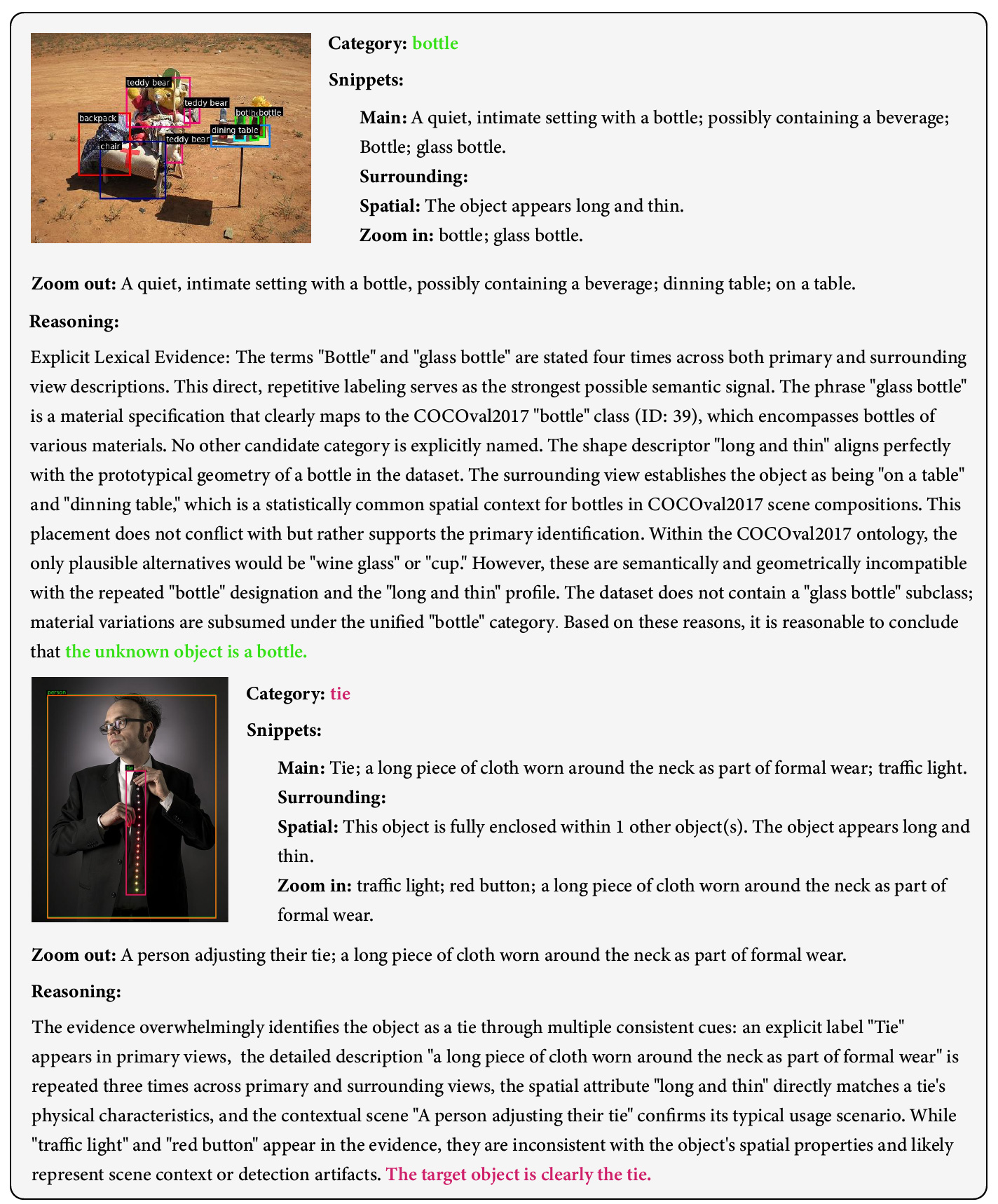}}
    \caption{LLM reasoning with snippet-based prompts to predict the categories ``bottle" and ``tie".}
    \label{fig:ns_1}
\end{figure*}

\begin{figure*}[!t]
    \centerline{\includegraphics[width=0.995\textwidth]{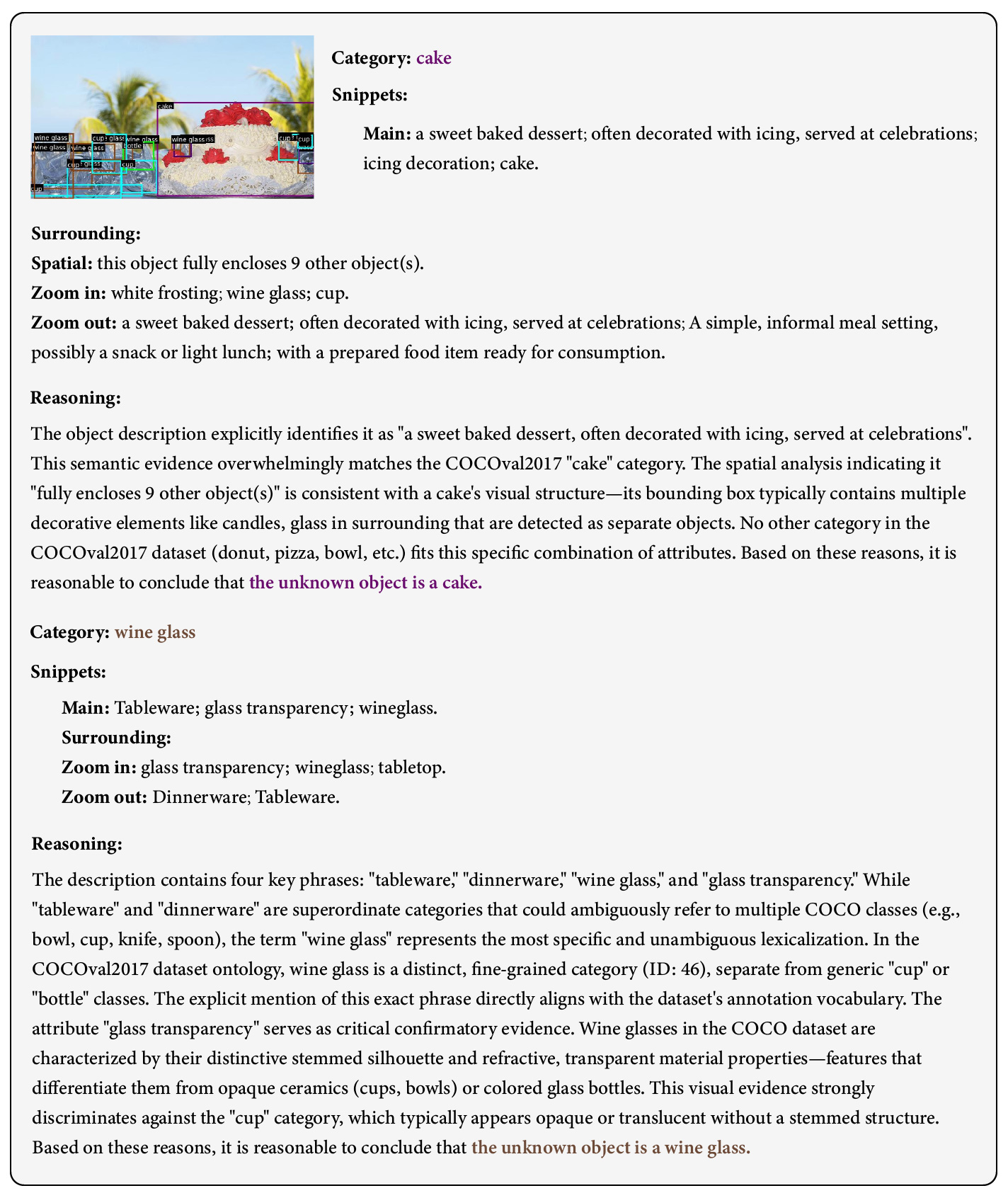}}
    \caption{LLM reasoning with snippet-based prompts to predict the categories ``cake" and ``wine glass".}
    \label{fig:ns_2}
\end{figure*}

\begin{figure*}[!t]
    \centerline{\includegraphics[width=0.995\textwidth]{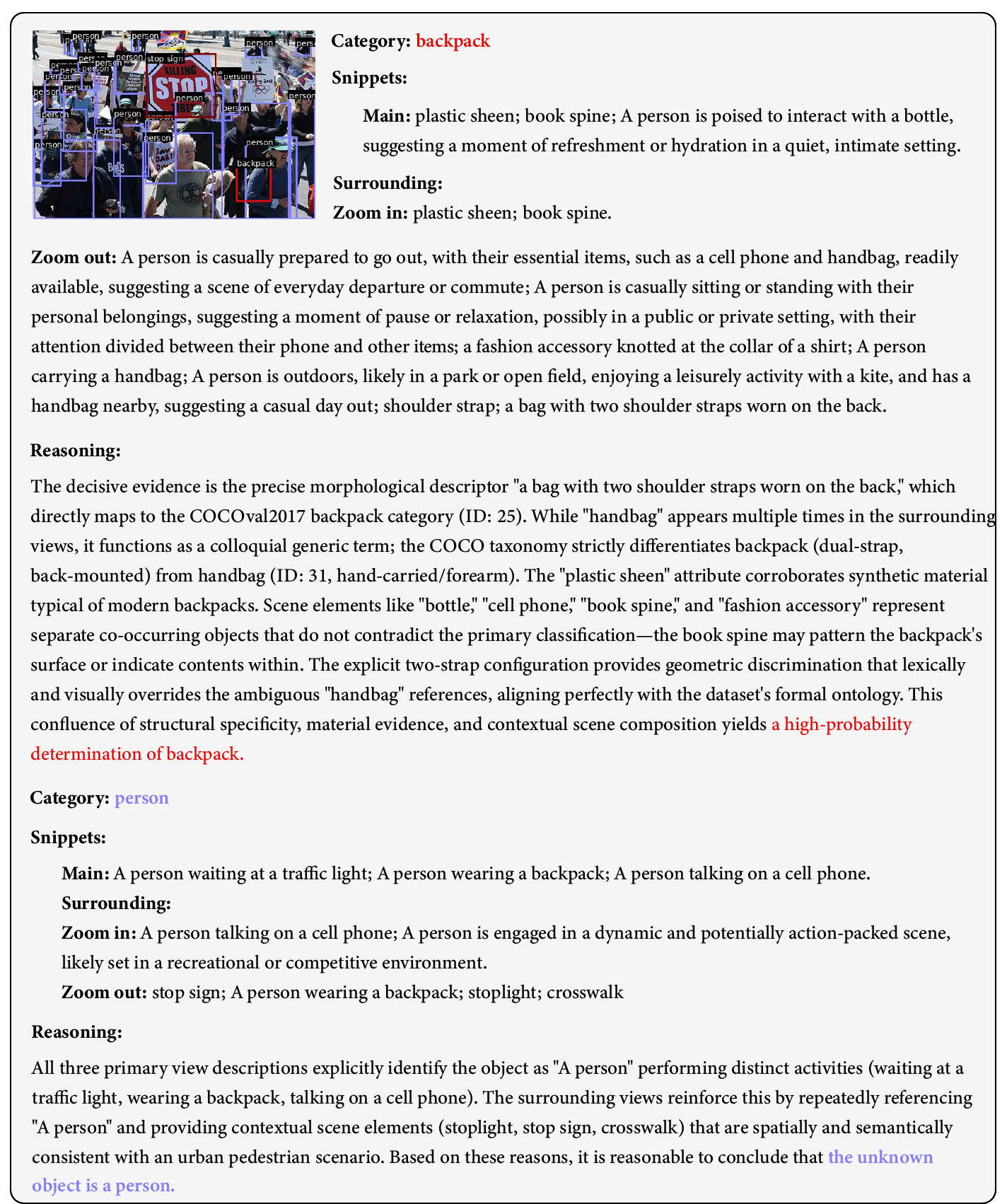}}
    \caption{LLM reasoning with snippet-based prompts to predict the categories ``backpack" and ``person".}
    \label{fig:ns_3}
\end{figure*}

\begin{figure*}[!t]
    \centerline{\includegraphics[width=0.995\textwidth]{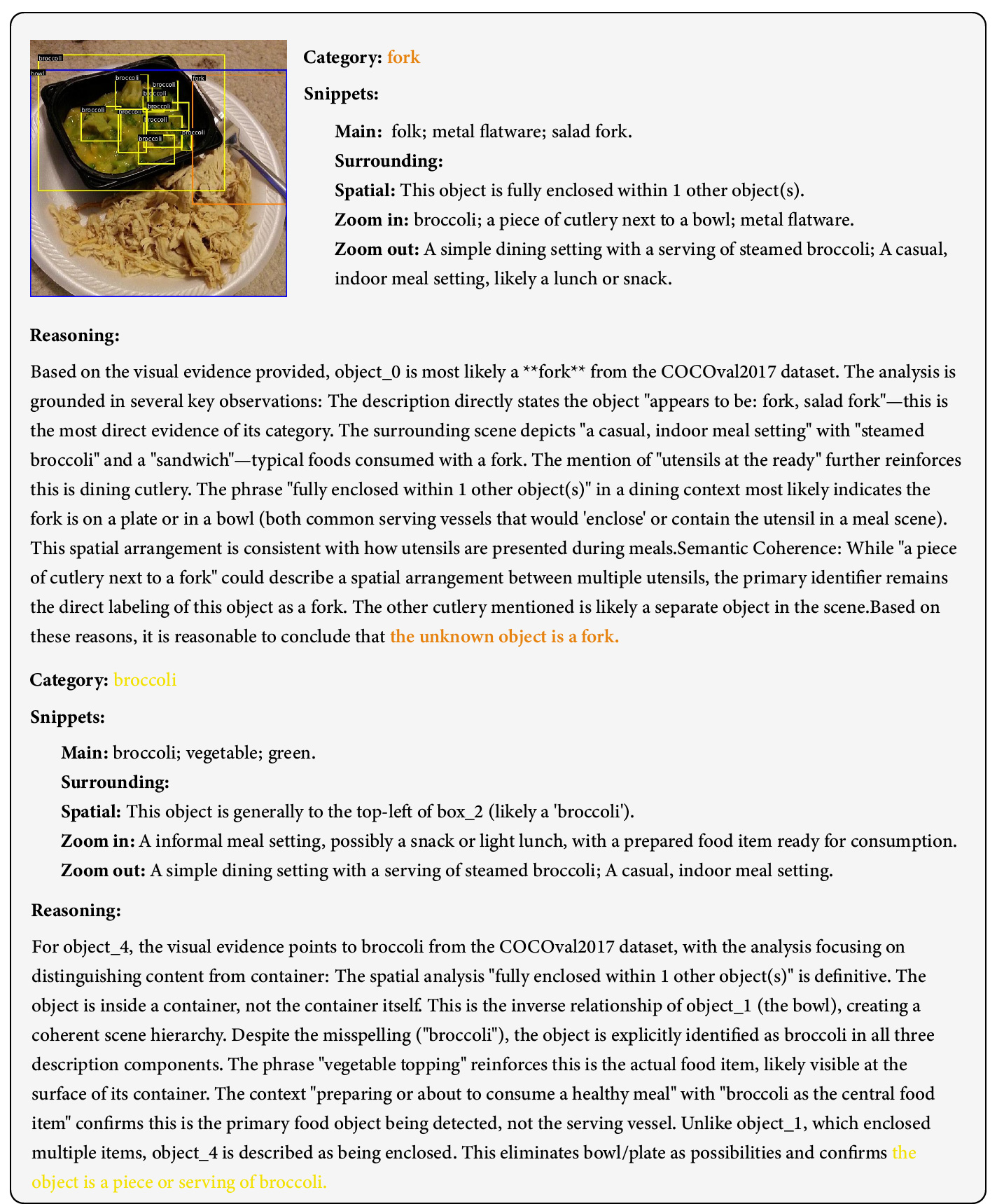}}
    \caption{LLM reasoning with snippet-based prompts to predict the categories ``fork" and ``broccoli".}
    \label{fig:ns_4}
\end{figure*}

\begin{figure*}[!t]
    \centerline{\includegraphics[width=0.995\textwidth]{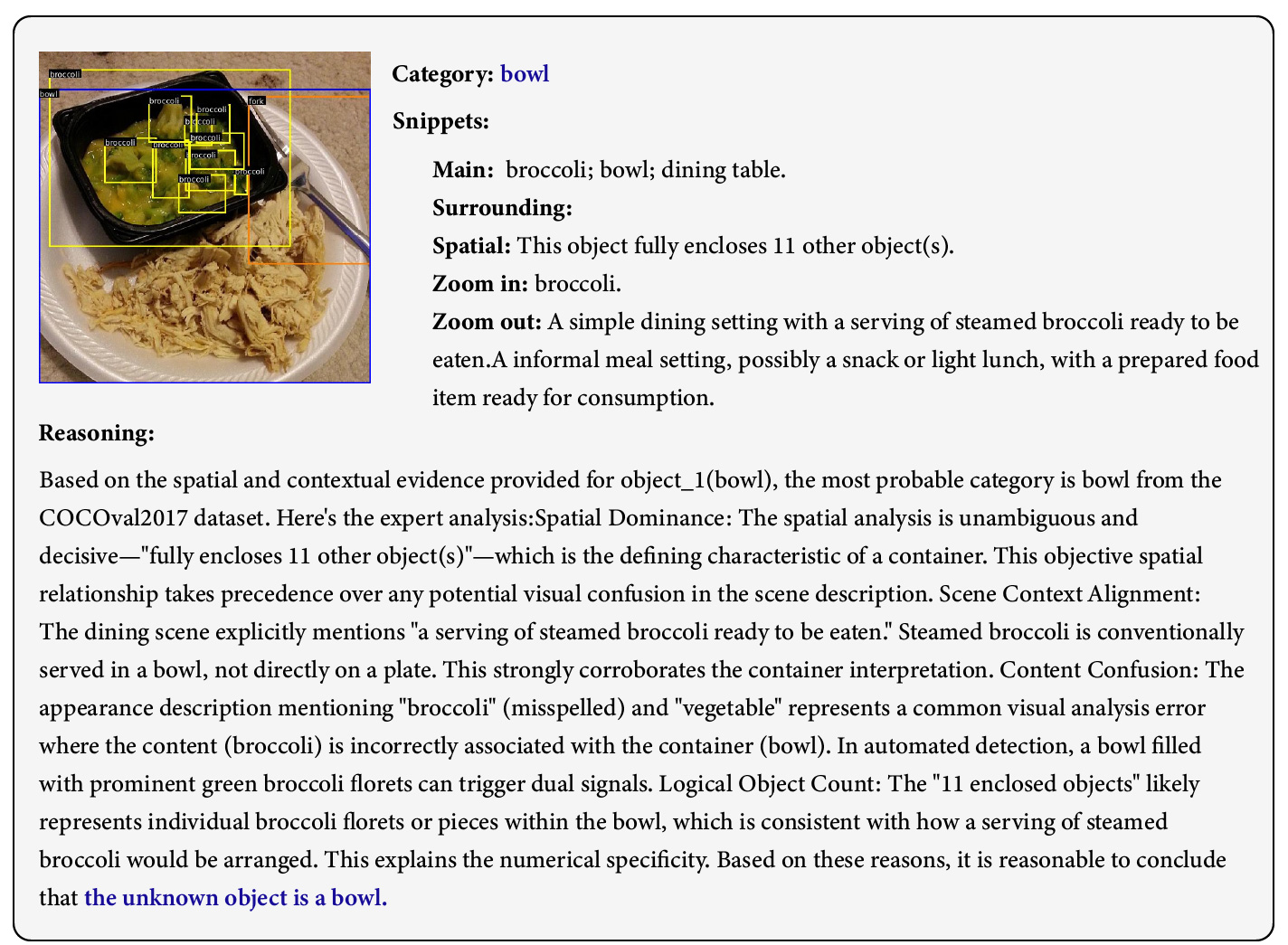}}
    \caption{LLM reasoning with snippet-based prompts to predict the category ``bowl".}
    \label{fig:ns_5}
\end{figure*}

This section provides a granular visualization of the model's inference trajectory, showcasing snippets from the designed Multi-Scale Visual Searching (MS-VLS), which includes a primary view for subject details and a surrounding view further stratified into Zoom-in and Zoom-out perspectives, alongside the corresponding LLM reasoning process. Here, these qualitative examples corroborate that the synergy between MS-VLS and the Contextual Concept Prompt (CCP) facilitates a comprehensive reasoning dynamic, enabling the LLM to effectively disambiguate confusing categories through contextual pipriors while identifying and suppressing irrelevant scene noise for robust classification.The results are presented in Figures~\ref{fig:rs_1}--\ref{fig:ns_5}.

Specific instances visually demonstrate these capabilities. As shown in Figure~\ref{fig:rs_2} and Figure~\ref{fig:rs_3}, the LLM exploits discriminative semantic cues, such as the \textit{three-point line} and \textit{wheels}, to successfully disambiguate \textit{basketball court} from \textit{tennis court}, and \textit{vehicle} from \textit{ship}, respectively. 

The noise suppression capability is exemplified in Figure~\ref{fig:ns_1}, where the LLM utilizes contextual information to explicitly identify \textit{traffic light} and \textit{red button} as environmental clutter unrelated to the target subject \textit{tie}, thereby ensuring precise classification. Furthermore, Figure~\ref{fig:ns_3} highlights the efficacy of multi-scale integration in understanding part-whole relationships; the model correctly interprets local details like \textit{plastic sheen} and \textit{book spine} as intrinsic surface attributes of a \textit{backpack}.

\begin{table*}[htbp]
  \centering
  \caption{Example generated snippets for both remote sensing and natural scenes.}
  \scalebox{0.83}{
  \begin{tabular}{c|>{\centering\arraybackslash}m{6.8cm}|c|>{\centering\arraybackslash}m{6.8cm}}
    \toprule
     Snippet Attributes & Remoteclip-snippets & Snippet Attributes & Siglip-snippets \\
    \midrule
    Appearance & Metal tanks with smooth, cylindrical surfaces arranged in rows; corrugated metal roof; smooth surface; paved surface; green grass; White or silver circular structure; streamlined; with multiple marked lines, including a three-point line... & High level categories & vehicle; animal; food; indoor item; tool; sports equipment; animal...  \\
    \midrule
    Shape & Rectangular shape; Circular shape; Star-shaped cross; Oval shape; Long straight line; Diamond shape; Tower shape; Building with a central dome; Lawn with circular pattern; pointed bow... & Common categories & person; bicycle; airplane; train; boat; cup; fork; knife; spoon; banana; apple...\\
    \midrule
    Relational & Ports are areas where ships dock to load/unload; container crane unloading/loading containers from ships; cargo box holds goods for transport; overpass above the highway; vehicles pass over other roads; roadway with rail tracks; long structure crossing a river/road/railway; wind turbine blades turned by wind; harbor with multiple boats docked; swimming pool complex with multiple pools... & Relational & A person riding a bicycle; A person driving a car; A person is seated on a bench; A small, tidy kitchen counter or windowsill with a touch of greenery; A person is engaged in a casual dining and work setup; A casual breakfast setting with a simple meal preparation...  \\
    \midrule
    Spatial & overpass above the highway; tennis court divided by a net; stadium with a large central field; raised structure with road running over another road; parking lot with parked trucks; Airport runway with multiple airplanes; highway with fast-moving traffic; bridge connecting two land masses... & Component attribute & wheel; handle; armrest;table leg; place mat; place setting; sharp edge; serrated edge; knife handle; two straps...  \\
    \midrule
    Semantic & \textbf{Transportation Infrastructure:} Ports; Dock; Container ship; Highway; Roundabout; Airport runway; Bicycle lane... \textbf{Industrial Areas:} Oil tanks; Storage units; Industrial or agricultural facility; Hydroelectric dam; Concrete dam; Gas station... \textbf{Sports Facilities:} Softball field; Baseball diamond; Tennis court; Large open-air stadium; Golf cart path; Ground track field... \textbf{Urban:} Architectural complex; Building with circular structure; Parking lot; Botanical garden; Concrete structure... \textbf{Natural Scenes:} Beach with forested area; Lake with boats; Garden or park; Mountains in the background; Beach with palm trees... & Scene description & A urban roadside setting with large vehicles and emergency infrastructure A person is engaged in personal mobile communication; A person is seated, likely taking a rest or pause, in a setting that appears to be a public or outdoor area given the presence of a bench; A person is sitting or standing near a fire hydrant, likely taking a break, with their personal belongings such as a handbag, a book, and a cell phone nearby, and they are also enjoying a beverage from a cup; A person is engaged in a healthy snack or meal, surrounded by fresh produce; A modern home office or entertainment space with a prominent workstation and multimedia setup, suggesting an environment where work, leisure, and possibly gaming or streaming activities converge; A controlled urban intersection where a vehicle has come to a halt A simple breakfast setting; A traveler preparing to venture out into potentially rainy weather...  \\
    \midrule
    Functional & ships dock to load/unload goods; runway used for planes takeoff/landing; dams control water flow/flood control; storage tanks store fuel/chemicals/oil; stadium used for large sports events; toll station used to charge vehicles; bridge used to cross obstacles... & Contextual clues & road surface; dining room setting; water surface; table setting; store aisle; shelf; supermarket; grocery store display; restaurant interior...  \\
  
    \bottomrule
  \end{tabular}
  }
  % \caption{}
  \label{tab:snippets} 
  
\end{table*}

\end{document}